\documentclass[runningheads]{llncs}
\usepackage{graphicx}
\usepackage{comment}
\usepackage{amsmath,amssymb} % define this before the line numbering.
\usepackage{color}

\usepackage{mathsymbols} % add mathmatical symbols

\usepackage{cite}
\usepackage{fixmath}
\usepackage{dsfont}
\usepackage{siunitx}
\usepackage{booktabs}
\usepackage{subfig}
\usepackage[font={footnotesize},skip=2pt]{caption}
\usepackage{floatrow}
\usepackage{xcolor}
\usepackage{hyperref}

\newfloatcommand{capbtabbox}{table}[][\FBwidth]

\newcommand\cbbest[1]{\setlength{\fboxsep}{0pt}\colorbox{gray!90}{\bf{#1}}}
\newcommand\cbsecond[1]{\setlength{\fboxsep}{0pt}\colorbox{gray!45}{\bf{#1}}}
\newcommand\cbthird[1]{\setlength{\fboxsep}{0pt}\colorbox{gray!20}{\bf{#1}}}

\makeatletter
\newcommand{\@chapapp}{\relax}%
\makeatother

\usepackage[toc,page]{appendix}

\begin{document}
%\renewcommand\thelinenumber{\color[rgb]{0.2,0.5,0.8}\normalfont\sffamily\scriptsize\arabic{linenumber}\color[rgb]{0,0,0}}
% \renewcommand\makeLineNumber {\hss\thelinenumber\ \hspace{6mm} \rlap{\hskip\textwidth\ 
%\hspace{6.5mm}\thelinenumber}}
% \linenumbers
\pagestyle{headings}
\mainmatter
\def\ECCVSubNumber{2754}  % Insert your submission number here

\title{DH3D: Deep Hierarchical 3D Descriptors for Robust
Large-Scale 6DoF Relocalization} % Replace with your title

\begin{comment}
\titlerunning{ECCV-20 submission ID \ECCVSubNumber}
\authorrunning{ECCV-20 submission ID \ECCVSubNumber}
\author{Anonymous ECCV submission}
\institute{Paper ID \ECCVSubNumber}
\end{comment}
%******************

%\begin{comment}
\titlerunning{DH3D: Deep Hierarchical 3D Descriptors for 6DoF Relocalization}
\author{Juan Du\inst{1 \star} \and
Rui Wang\inst{1,2}\thanks{Authors contributed equally.} \and
Daniel Cremers\inst{1,2}}
\authorrunning{J. Du et al.}
\institute{Technical University of Munich \and
Artisense\\
\email{\{duj, wangr, cremers\}@in.tum.de}}
%\end{comment}
%******************
\maketitle

\begin{abstract}
For relocalization in large-scale point clouds, we propose the first approach that 
unifies global place recognition and local 6DoF pose refinement. To this end, we design a Siamese 
network that jointly learns 3D local feature detection and description directly from raw 3D points. 
It integrates FlexConv and Squeeze-and-Excitation (SE) to assure that the learned local descriptor 
captures multi-level geometric information and channel-wise relations. For detecting 3D keypoints 
we predict the discriminativeness of the local descriptors in an unsupervised manner. We generate 
the global descriptor by directly aggregating the learned local descriptors with an effective 
attention mechanism. In this way, local and global 3D descriptors are inferred in one single 
forward pass. Experiments on various benchmarks demonstrate that our method achieves competitive 
results for both global point cloud retrieval and local point cloud registration in 
comparison to state-of-the-art approaches. To validate the generalizability and robustness of our 
3D keypoints, we demonstrate that our method also performs favorably without fine-tuning on 
the registration of point clouds that were generated by a visual SLAM system. Code and related 
materials are available at \url{https://vision.in.tum.de/research/vslam/dh3d}.
\end{abstract}

\section{Introduction}
\begin{figure}[htbp]
	\centering
	\includegraphics[width=0.98\textwidth]{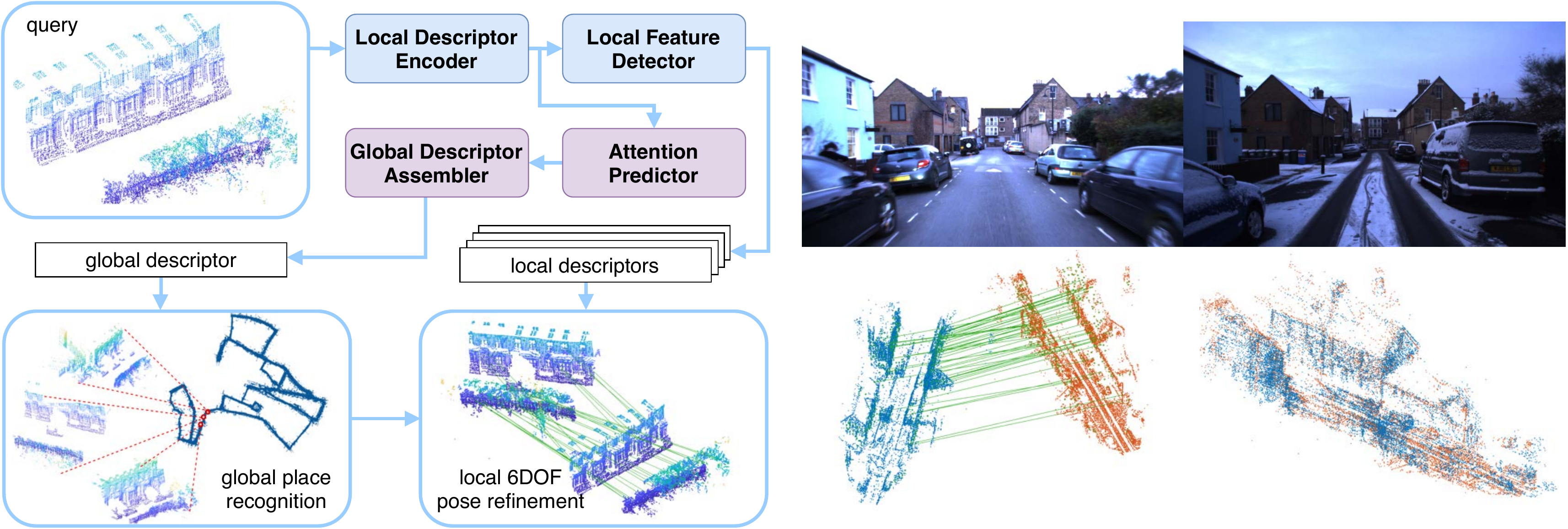} %%
	\caption{Left: We propose a hierarchical network for large-scale point cloud based 
	relocalization. The network	consumes raw 3D points and performs local feature detection, 
	description and global descriptor extraction in one forward pass. The global descriptor is used 
	to retrieve similar scenes from the database. Accurate 6DoF pose is then obtained by matching 
	the local features. Right: Our local descriptors trained on	LiDAR points work favorably on 
	the sparse point clouds generated by a visual SLAM method without fine-tuning. The point clouds 
	are generated from sequences with different weathers, lighting conditions and scene layouts, 
	thus \textit{have significantly different distributions}.}
	\label{fig:teaser}
\end{figure}

Relocalization within an existing 3D map is a critical functionality for numerous applications in 
robotics~\cite{axelrod2018provably} and autonomous driving~\cite{wang2018dels, ort2018autonomous}. 
A common strategy is to split the problem into two subtasks, namely global place recognition and 
local 6DoF pose refinement. A lot of effort has been focused on tackling the problem using 2D
images~\cite{sarlin2019coarse,taira2018inloc, chen2017deep,	sattler2017large}, where the 3D maps 
are usually defined as image feature points reconstructed in 3D using Structure from Motion (SfM). 
The coarse global
place recognition is achieved by image retrieval, whereas accurate local 6DoF pose refinement is 
addressed separately by feature matching and PnP. With the progress of deep learning for image 
descriptor extraction~\cite{netvlad, gordo2016deep} and 2D keypoints 
detection/description~\cite{yi2016lift, detone2018superpoint, d2net, r2d2}, image based methods 
have significantly gained in robustness to variations in viewpoint and illumination.

Alternatively one can tackle these variations by working on 3D point clouds 
since these are inherently invariant to such issues. Moreover, there exist numerous SLAM 
pipelines that generate accurate large-scale point clouds using sensory input from 
LiDAR~\cite{Zhang2014RSS, zhang2015visual, deschaud2018imls} or camera~\cite{engel2015large, 
stereodso}. While there is great potential to rely on such data, research on point cloud based
relocalization is significantly less matured compared to the image-based 
counterpart~\cite{r2d2, d2net}. Especially, deep learning on 3D descriptors emerged only roughly 3 
years ago. With most of the early attempts focusing on small-scale tasks like object 
classification, detection and segmentation~\cite{pointnet, pointnetplus, pointcnn, 
zhou2018voxelnet, instanceseg}, only a limited number of networks have been proposed for 
large-scale localization~\cite{lu2019l3, lu2019deepvcp, wang2019deep}. Moreover, among these few 
attempts, global place recognition~\cite{pointnetvlad, pcan} and local 6DoF pose 
refinement~\cite{3dfeat, gojcic2019perfect, choy2019fully} have been addressed isolatedly, despite 
the fact that both tasks depend on the same low level geometric clues. 

In this paper, we propose a hierarchical deep network for large-scale point clouds based 
relocalization -- see Fig.~\ref{fig:teaser}. The network directly consumes unordered 3D points and 
performs keypoint detection and description, as well as global point cloud descriptor extraction 
in a unified manner. In contrast to the conventional \textit{detect-then-describe} pipeline, our 
local features are learned with the \textit{detect-and-describe} concept. We estimate a confidence 
map of the discriminativeness of local features explicitly and learn to select keypoints that are 
well-suited for matching in an unsupervised manner. The local features are aggregated into a global 
descriptor for global retrieval, attaining a consistent workflow for large-scale outdoor 6DoF 
relocalization. Our main contributions are summarized as follows:
\begin{itemize}
	\small
	\itemsep0em
	\item We propose the first work that unifies point cloud based global place recognition and 
	6DoF pose refinement. Our method performs local feature detection and description, as well as 
	global descriptor extraction in one forward pass, running significantly faster than 
	previous methods. 
	\item We propose to use FlexConv and SE block to integrate multi-level context information and 
	channel-wise relations into the local features, thus achieve much stronger performance on 
	feature matching and also boost the global descriptor.
	\item We introduce a \textit{describe-and-detect} approach to explicitly learn a 3D keypoint 
	detector in an unsupervised manner.
	\item Both our local and global descriptors achieve state-of-the-art performances on point 
	cloud registration and retrieval across multiple benchmarks.
	\item Furthermore, our local descriptors trained on LiDAR data show competitive generalization 
	capability when applied to the point clouds generated by a visual SLAM method, even though 
	LiDAR and visual SLAM point clouds exhibit very different patterns and distributions.
\end{itemize}

\section{Related Work}
\medskip
\noindent
%-----------------------------------------------------------------------------------------------
{\bf Handcrafted local descriptors} encode local structural information as histograms over 
geometric properties e.g., surface normals and curvatures. Spin image (SI)~\cite{johnson1997spin} 
projects 3D points within a cylinder onto a 2D spin image. Unique Shape Context 
(USC)~\cite{tombari2010unique} deploys a unique local reference frame to improve the accuracy of 
the well-know 3D shape context descriptor. Point Feature Histogram (PFH)~\cite{rusu2008persistent} 
and Fast PFH (FPFH)~\cite{rusu2010fast} describe the relationships between a point and its 
neighbors by calculating the angular features and normals. While these handcrafted methods have 
made great progress, they generalize poorly to large-scale scenarios and struggle to handle noisy 
real-world data.

\medskip
\noindent
%----------------------------------------------------------------------------------------------
{\bf Learned local descriptors.}  
To cope with the inherent irregularity of point cloud data, researchers have suggested to convert
3D points to regular representations such as voxels and multi-view 3D images~\cite{multiview,  
octreecnn, qi2016volumetric, dynamic, splatnet}. As a pioneering work, PointNet~\cite{pointnet} 
proposed to apply deep networks directly to raw 3D points. Since then, new 
models~\cite{pointnetplus, pointcnn, zhao20193d} and flexible operators on irregular 
data~\cite{flex, wang2018deep, wang2019dynamic} have been emerging. 
Accompanied by these progresses, learning-based 3D local descriptors such as 
3DMatch~\cite{3dmatch}, PPFNet~\cite{ppfnet} and PPF-FoldNet~\cite{deng2018ppf, deng20193d} have 
been proposed for segment matching, yet they are designed for RGB-D based indoor applications. 
Recently, Fully Convolutional Geometric Features (FCGF)~\cite{choy2019fully} was proposed to 
extract geometric features from 3D points. Yet, all these methods do not tackle feature detection 
and get their features rather by sampling. Another class of methods utilizes deep learning to 
reduce the dimensions of the handcrafted descriptors, such as Compact Geometric Features 
(CGF)~\cite{khoury2017learning} and LORAX~\cite{elbaz20173d}. 
In the realm of large-scale outdoor relocalization, 3DFeatNet~\cite{3dfeat} and 
L\textsuperscript{3}-Net~\cite{lu2019l3} extract local feature embedding using PointNet, whereas 
3DSmoothNet~\cite{gojcic2019perfect} and DeepVCP~\cite{lu2019deepvcp} rely on 3D CNNs. In contrary 
to registration based on feature matching, DeepVCP　and Deep Closest Point~\cite{wang2019deep} learn 
to locate the correspondences in the target point clouds.

\medskip
\noindent
%-----------------------------------------------------------------------------------------------
{\bf 3D keypoint detectors.}
There are three representative hand-crafted 3D detectors. Intrinsic Shape Signatures 
(ISS)~\cite{iss3d} selects salient points with large variations along the principal axes. 
SIFT-3D~\cite{sift3d} constructs a scale-space of the curvature with the DoG operator. 
Harris-3D~\cite{sipiran2011harris} calculates the Harris response of each 3D vertex based on first 
order derivatives along two orthogonal directions on the 3D surface. Despite the increasing 
number of learning-based 3D descriptors, only a few methods have been proposed to learn to detect 
3D keypoints. 3DFeatNet~\cite{3dfeat} and DeepVCP~\cite{lu2019deepvcp} use an attention layer to 
learn to weigh the local descriptors in their loss functions. Recently, USIP~\cite{li2019usip} has 
been proposed to specifically detect keypoints with high repeatability and accurate localization. 
It establishes the current state of the art for 3D keypoint detectors.

\medskip
\noindent
%----------------------------------------------------------------------------------------------
\textbf{Handcrafted global descriptors.}
Most 3D global descriptors describe places with handcrafted statistical information. Rohling et 
al.~\cite{rohling2015fast} propose to describe places by histograms of points elevation. Cop et 
al.~\cite{cop2018delight} leverage LiDAR intensities and present DELIGHT. Cao et 
al.~\cite{cao2018robust} transform a point cloud to a bearing-angle image and extract ORB features 
for bag-of-words aggregation.

\medskip
\noindent
%-----------------------------------------------------------------------------------------------
\textbf{Learned global descriptors.} 
Granstr{\"o}m et al.~\cite{granstrom2011learning} describe point clouds with rotation invariant 
features and input them to a learning-based classifier for matching. LocNet~\cite{yin2018locnet} 
inputs range histogram features to 2D CNNs to learn a descriptor. In these methods, deep learning 
essentially plays the role of post-processing the handcrafted descriptors. Kim et 
al.~\cite{kim2018scan} transform point clouds into scan context images and feed them into CNNs for 
place recognition. 
PointNetVLAD~\cite{pointnetvlad} first tackles place recognition in an end-to-end way. The global 
descriptor is computed by a NetVLAD~\cite{netvlad} layer on top of the feature map extracted using 
PointNet~\cite{pointnet}. Following this, PCAN~\cite{pcan} learns attentions for points to produce 
more discriminative descriptors. These two methods extract local features using PointNet, which 
projects each point independently into a higher dimension and thus does not explicitly use 
contextual information.

\section{Hierarchical 3D Descriptors Learning}
For large-scale relocalization, an intuitive approach is to tackle the problem hierarchically in a 
coarse-to-fine manner: local descriptors are extracted, aggregated into global 
descriptors for coarse place recognition, and then re-used for accurate 6DoF pose refinement. While 
being widely adopted in image the domain~\cite{orbslam, sarlin2018leveraging, sarlin2019coarse}, 
this idea has not been addressed by the deep learning community for 3D. As a result, seeking for 
6DoF relocalization for point clouds, one has to perform local feature detection, description, 
global descriptor extraction separately, possibly running an independent network for each. To 
address this problem, we design a hierarchical network operating directly on a point cloud, 
delivering local descriptors, a keypoint score map and a global descriptor in a single forward 
pass. Point cloud based relocalization thus can be performed hierarchically: a coarse search using 
the global descriptor retrieves 3D submap candidates, which are subsequently verified by local 3D 
feature matching to estimate the 6DoF poses. An overview of our system is provided in 
Fig.~\ref{fig:teaser}.

\subsection{3D Local Feature Encoder and Detector}
3DFeatNet~\cite{3dfeat} is a seminal work that learns both 3D local feature detection and 
description. 
Nevertheless, the following two points potentially limit its discriminative power: (1) Its detector 
is an attention map learned directly from the input points. 
During inference descriptors are only extracted for the keypoints defined by the attention map. 
Such classical \textit{detect-then-describe} approach, as discussed in~\cite{d2net,r2d2}, typically 
focuses on low-level structures of the raw input data, and cannot utilize the high level 
information encoded in the descriptors. (2) Its feature description is PointNet-based, the 
symmetric function of which tends to provide only limited structural information of local clusters. 
To resolve these limitations, we propose to use Flex Convolution 
(FlexConv)~\cite{flex} and Squeeze-and-Excitation (SE) block~\cite{hu2018squeeze} to respectively 
fuse multi-level spatial contextual information and channel-wise feature correlations into the 
local descriptors. The \textit{describe-and-detect} pipeline~\cite{d2net,r2d2} is adopted to 
postpone our detection stage to employ higher-level information in the learned descriptors.
\begin{figure}[t]
	\centering
	\includegraphics[width=0.95\textwidth]{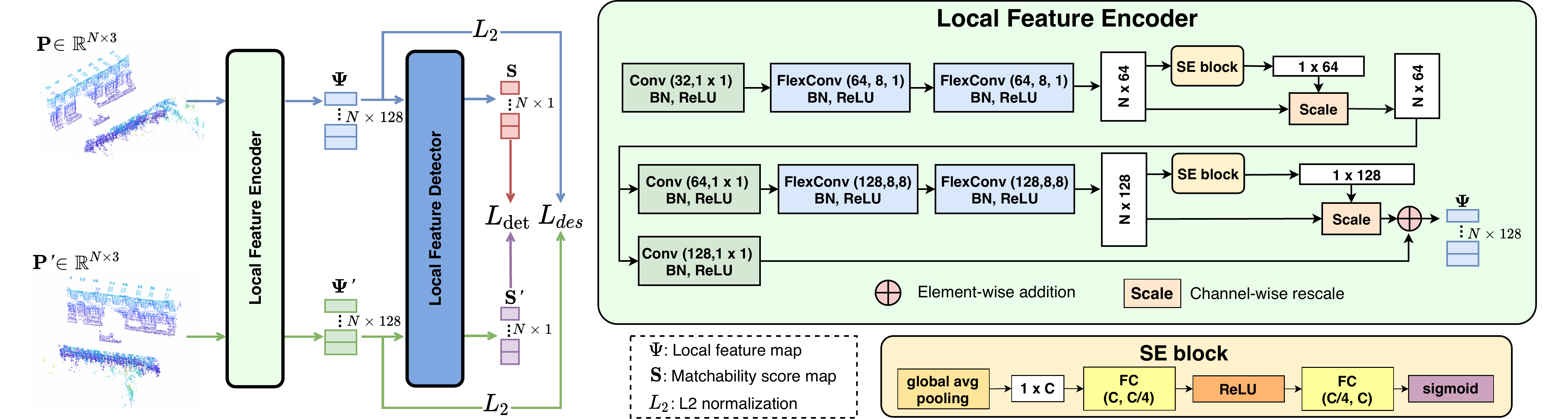}
	\caption{Left: the flow of local feature descriptor and detector learning. Right: the 
		architecture of our Local Feature Encoder. In Conv($D$, $K\times K$) and FlexConv($D$, $k$, 
		$d$), $D$: output dimension, $K$: filter size, $k$: neighborhood size, $d$: dilation 
		rate.}
	\label{fig:local_net}
\end{figure}

\medskip
\noindent
%-----------------------------------------------------------------------------------------------
{\bf FlexConv layer.}
Considering a 3D point \(\vp_l\) with its \(k\) neighbors \(N_k(\vp_l) = \lbrace{\vp_{l_1},\cdots, 
\vp_{l_k}}\rbrace\) as a graph $\mathcal{G} = (\mathcal{V}, \mathcal{E})$, where 
$\mathcal{V}=\{0,..,k\}$ are the vertices and $\mathcal{E} \subseteq \mathcal{V}\times\mathcal{V}$ 
the edges. A 3D operator on an edge of this graph can be formulated as 
$\mathbf{e}_{ll_{k}}=f_{\mathbf{\Theta}}(\vp_l, \vp_{l_k})$, with $\mathbf{\Theta}$ the set of 
learnable parameters. As pointed out by~\cite{wang2019dynamic}, PointNet is a special case of 
$f_{\mathbf{\Theta}}(\vp_l, \vp_{l_k}) = f_{\mathbf{\Theta}}(\vp_l)$, thus encodes only global 
shape information and ignores the local neighborhood structure. In contrary, FlexConv can be 
abstracted as $f_{\mathbf{\Theta}}(\vp_l, \vp_{l_k}) = f_{\mathbf{\Theta}}(\vp_l - \vp_{l_k}, 
\vp_{l_k})$, therefore can effectively encode local information, which we believe is crucial for 
learning discriminative local descriptors. Formally, FlexConv is a generalization of the 
conventional grid-based convolution and is defined as:
\begin{equation} \label{eq:flexconv}
{f_{FlexConv}}(\vp_l) = \sum_{\vp_{l_i}\in N_k(\vp_l)}\omega(
\vp_{l_i}, \vp_l) \cdot h(\vp_{l_i}),
\end{equation}
where \(h(\vp_{l_i}) \in \mathbb{R}^C\) is a point-wise encoding function projecting a point to 
the high-dimensional feature space. It is convolved with a filter-kernel \(\omega: \mathbb{R}^3 
\times \mathbb{R}^3 \rightarrow \mathbb{R}^C\) that is computed by the standard scalar product in 
the Euclidean space, with learnable parameters \({\theta} \in \mathbb{R}^{C\times 3}\), 
\({\theta_{b}} \in \mathbb{R}^{C}: w(\vp_l, \vp_{l_i} \mid {\theta}, {\theta_b}) = \langle \theta, 
\vp_l - \vp_{l_i}\rangle + \theta_{b}\). It can be considered as a linear approximation of the 
traditional filter-kernel which uses the location information explicitly. In addition, \(\omega\) 
is everywhere well-defined and allows us to perform back-propagation easily.

\medskip
\noindent
{\bf Squeeze-and-Excitation (SE) block.}
While FlexConv models spatial connectivity patterns, SE blocks~\cite{hu2018squeeze} are further 
used to explicitly model the channel-wise inter-dependencies of the output features from the 
FlexConv layers. Let \(\mathbold{U} = \lbrace u_1, \cdots, u_C \rbrace\in \mathbb{R}^{N\times C}\) 
denote the input feature map to the SE block, where \(u_c \in \mathbb{R}^N\) represents the 
\(c{\text{-th}}\) channel vector of the output from the last FlexConv layer. The squeeze operation 
first ``squeezes'' \(\mathbold{U}\) into a channel-wise descriptor \(z \in \mathbb{R}^{C}\) as 
$f_{sq}: \mathbb{R}^{N\times C} \rightarrow \mathbb{R}^C, \quad z = f_{sq} (\mathbold{U})$,
where \(f_{sq}\) is implemented as a global average pooling to aggregate across spatial 
dimensions. \(z \in \mathbb{R}^{C}\) is an embedding containing global information which is then 
processed by the excitation operation $f_{ex}: \mathbb{R}^{C} \rightarrow \mathbb{R}^C, \quad s = 
f_{ex} (z)$, 
where \(s \in \mathbb{R}^{C} \) is implemented as two fully connected layers with ReLU to fully 
capture channel-wise dependencies and to learn a nonlinear relationship between the channels. In 
the end, the learned channel activations are used to recalibrate the input across channels 
achieving the attention selection of different channel-wise features $\tilde{u}_c = f_{scale} (u_c, 
s_c) = s_c \cdot u_c$, where \(\tilde{U} = \lbrace \tilde{u}_1, \cdots, \tilde{u}_C\rbrace\) refers 
to the output of the SE block.

\medskip
\noindent
%-----------------------------------------------------------------------------------------------
{\bf Encoder architecture.}
The architecture of the encoder module is illustrated in Fig.~\ref{fig:local_net}. In comparison to 
3DFeatNet~\cite{3dfeat} which relies on PointNet and only operates on one level of spatial 
granularity, our 
encoder extracts structural information from two spatial resolutions. At each resolution, the 
following operations are conducted: one \(1\times1\) convolution, two consecutive FlexConv layers 
and a SE block. Taking a point cloud \(\mathbold{P} = \lbrace\vp_{1},\cdots,\vp_{N}\rbrace \in 
\mathbb{R}^{N \times 3}\) as input, the encoder fuses multi-level contextual information 
by adding the outputs from the two resolutions and produces the feature map \(\mathbold{\Psi}\). It 
is then L2-normalized to give us the final local descriptor map \(\goodchi =  
\lbrace\vx_{1},\cdots,\vx_{N}\rbrace \in  \mathbb{R}^{N \times D}\). 
Benefited from the better policies for integrating contextual information, when compared to 
3DFeatNet, our local features are much more robust to points densities and distributions, thus 
generalize significantly better to point clouds generated by different sensors (more details in 
Sec.~\ref{app_SLAM}).

\medskip
\noindent
%-----------------------------------------------------------------------------------------------
{\bf Description loss.}
In feature space, descriptors of positive pairs are expected to be close and those of negative 
pairs should keep enough separability. Instead of using the simple Triplet Loss as in 3DFeatNet, we 
adopt the N-tuple loss~\cite{ppfnet} to learn to differentiate as many patches as possible.	
Formally, given two point clouds 
\(\mat{P}\), \(\mat{P}'\), the two following matrices can be computed: a feature space distance 
matrix \(\mat{D}\ \in \realset^{N\times N} \) with \(\mat{D}(i,j) = \|\vec{x}_i - \vec{x}_j \|\), a 
correspondence matrix \(\gt \in \realset^{N\times N}\) with \(\gt_{i,j}\in\{0,1\}\) indicating 
whether 
two point patches \(\vp_{i} \in \mat{P}\) and \(\vp_{j} \in \mat{P}'\) form a positive pair, i.e., 
if their distance is within a pre-defined threshold. The N-tuple loss is formulated as: 
\begin{equation} \label{eq:contrastive2}
L_{desc} = \sideset{}{^*}\sum{ \biggl(\frac{\gt \circ \mat{D}}{\|\gt\|^2_F} + \eta \frac{max(\mu - 
(1-\gt)\circ \mat{D}, 0)}{N^2 - \|\gt\|^2_F}\biggr) },
\end{equation}
where \(\sum^*(\cdot)\) is element-wise sum, \(\circ\) element-wise multiplication, $\|\cdot\|_F$ 
the Frobenius norm, \(\eta\) a hyper-parameter balancing matching and non-matching pairs. 
The loss is divided by the number of true/false matches to remove the bias introduced by the larger 
number of negatives.

\medskip
\noindent
%-----------------------------------------------------------------------------------------------
{\bf 3D local feature detection.}
Contrary to the classical \textit{detect-then-describe} approaches, we postpone the detection to a 
later stage. To this end, we produce a keypoint saliency map \(\mathbold{S} \in 
\mathbb{R}^{N\times1}\) from the extracted point-wise descriptors instead of from the raw input 
points. Our saliency map is thus estimated based on the learned local structure encoding 
and thus is less fragile to low level artifacts in the raw data, and provides significantly better 
generalization. Another benefit of the \textit{describe-and-detect} pipeline is that feature 
description and detection can be performed in one forward pass, unlike the 
\textit{detect-then-describe} approaches that usually need two stages. Our detector consumes the 
local feature map \(\mathbold{\Psi}\) by a series of four \(1\times1\) convolution layers, 
terminated by the sigmoid activation function (more details in the supplementary document). 

As there is no standard definition of a keypoint's discriminativeness for outdoor point clouds, 
keypoint detection cannot be addressed by supervised learning. In our case, since the learned 
descriptors are to be used for point cloud registration, we propose to optimize keypoint 
confidences by leveraging the quality of descriptor matching. Local descriptor matching essentially 
boils down to nearest neighbor search in the feature space. Assuming the descriptor is informative 
enough and the existence of correspondence is guaranteed, a reliable keypoint, i.e., a keypoint 
with a high score \(s\), is expected to find the correct match with high probability. We therefore 
can measure the quality of the learned detector using $\eta_i = (1-s_i)\cdot (1-\gt_{i,j}) + 
s_i\cdot \gt_{i,j}$, where \(s_i \in [0,1]\) is an element of \(\lmap\) and \(j\) refers to the 
nearest neighbor in \(\mathbold{\Psi}'\). A simple loss function can be formulated as $L_{det} = 
\frac{1}{N}\sum_{i=1}^{N} 1 - \eta_i$. However, we find that only using the nearest neighbor to 
define \(\eta\) is too strict on the learned feature quality and the training can be unstable. We 
thus propose a new metric called average successful rate (\(\ar\)): given a point \(\vp_i \in 
\mat{P}\) and its feature \(\mathbold{\psi}_i \in \mathbold{\Psi}\), we find the \(k\) nearest 
neighbors in \(\mathbold{\Psi}'\). The \(\ar\) of \(\vp_i\) is computed as: $\ar_i =\frac{1}{k} 
\sum_{j=1}^{k} c_{ij}$, where \(c_{ij} = 1\) if at least one correct correspondence 
can be found in the first \(j\) candidates, otherwise is 0.\footnote{E.g., if the first correct 
correspondence appears as the 3rd nearest neighbor, then \(\ar\) in the case of \(k=5\) is \((0 + 0 
+ 1 + 1 + 1) / 5 = 0.6\).} Now we can measure \(\eta\) with \(\ar\) which is a real number in the 
range \([0,1]\) instead of a binary number and the loss above can be rewritten as: 
\begin{equation} \label{eq:detloss1}
L_{det} = \frac{1}{N}\sum_{i=1}^{N} 1 - [ \kappa(1-s_i) + s_i \cdot \ar_i ], 
\end{equation}
where \(\kappa \in [0,1]\) is a hyperparameter indicating the minimum expected \(\ar\) per
keypoint. To minimize the new loss, the network should predict \(s_i\) to be close to 0 if \(\ar_i 
< \kappa\) and to be near 1 conversely.

\subsection{Global Descriptor Learning}
As a key concept of this work, we propose to re-use the local descriptors for global retrieval. 
This early sharing of the computation is natural as both local and global descriptors are based on 
the same low-level geometric clues. The upcoming question is, how can the local descriptors be 
aggregated to a global one? While there exist many ways to do so, e.g., pooling, 
PointNet++~\cite{pointnetplus}, FlexConv, PointCNN~\cite{pointcnn}, Dynamic Graph 
CNN~\cite{wang2019dynamic}, we claim that the PCAN~\cite{pcan} (PointNetVLAD extended by adding 
attention) gives the best performance among the many and provide an ablation 
study in the supplementary material.

\begin{figure}[t]
	\centering
	\includegraphics[width=0.9\textwidth]{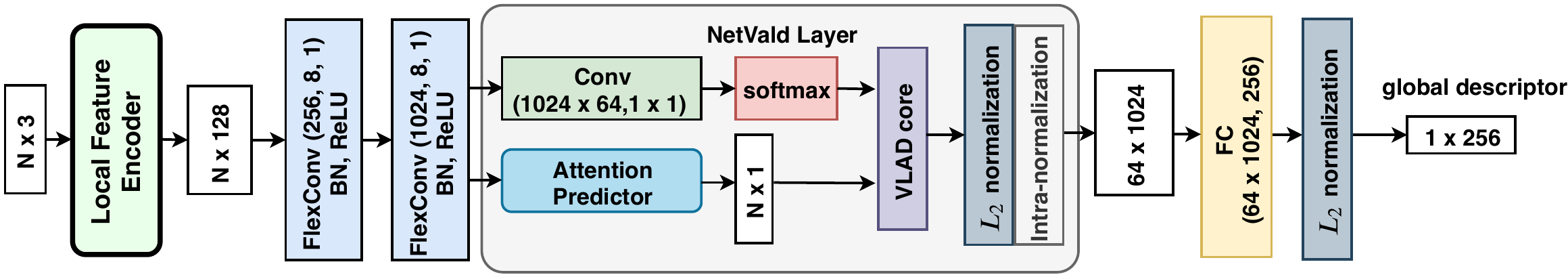}
	\caption{ The architecture of the global descriptor assembler.}
	\label{fig:global_net}
\end{figure}

Our global aggregation network is depicted in Fig.~\ref{fig:global_net}. 
Before the NetVLAD module, two FlexConv layers are added to project the local features to a higher 
dimension for a more retrieval relevant encoding. The attention predictor takes these features and 
outputs a per-point attention map, which is followed by a NetVLAD layer to generate a compact 
global representation. As the output of a NetVLAD layer usually has very high dimension which 
indicates expensive nearest neighbor search, a FC layer is used to compress it into a
lower dimension. The global descriptor assembler is trained using the same lazy quadruplet loss as 
used in~\cite{pointnetvlad, pcan}, to better verify our idea of using the learned local descriptor 
for global descriptor aggregation.

\medskip
\noindent
%-----------------------------------------------------------------------------------------------
{\bf Attention map prediction.}
As it has been observed for image retrieval, visual cues relevant to place recognition are 
generally not uniformly distributed across an image. Therefore focusing on important regions is the 
key to improve the performance~\cite{noh2017large,chen2018learning}. However, such attention 
mechanism has only been explored recently for point cloud retrieval. 
Inspired by PCAN~\cite{pcan}, we integrate an attention module that weighs each local descriptor 
before aggregation. As a key difference to PCAN, the input to our attention predictor is the 
learned descriptors which already encapsulate fairly good contextual information. Thus our 
predictor is not in charge of aggregating neighborhood information and needs a dedicated design to 
reflect such benefit. We thus construct our attention predictor by only chaining up three 1x1 Conv 
layers followed by softmax to ensure the  sum of the attention weights is 1. We will show that, 
although our attention predictor has a much simpler structure than PCAN, yet it is effective. When 
combined with our descriptor, it still offers better global retrieval performance.
More details on the network structure are provided in the supplementary document.

%%%%%%%%%%%%%%%%%%%%%%%%%%%%%%%%%%%%%%%%%%%%%%%%%%%%%%%%%%%%%%%%%%%%%%%%%%%%%%%%%%%%%%%%%%%%%%%%%%%%
\section{Experiments}
%%%%%%%%%%%%%%%%%%%%%%%%%%%%%%%%%%%%%%%%%%%%%%%%%%%%%%%%%%%%%%%%%%%%%%%%%%%%%%%%%%%%%%%%%%%%%%%%%%%%
The LiDAR point clouds from the Oxford RobotCar dataset~\cite{robotcar} are used to train our 
network. Additionally, the ETH dataset~\cite{pomerleau2012challenging} and the point clouds 
generated by Stereo DSO~\cite{stereodso}, a direct visual SLAM method are used to test the 
generalization ability of the evaluated methods. The margin used in \(L_{desc}\) is set to \(\mu = 
0.5\), the minimum expected \(\ar\) in Eq.~\ref{eq:detloss1} \(\kappa=0.6\) with \(k=5\), \(N\) in 
Eq.~\ref{eq:contrastive2} is set to 512. We use a weighted sum of \(L_{desc}\) and \(L_{det}\) as 
the loss function to train our network $L = L_{desc} + \lambda L_{det}$.

To train the local part of our network, we use Oxford RobotCar and follow the data processing 
procedures in~\cite{3dfeat}. We use 35 traversals and for each create 3D submaps using the provided 
GPS/INS poses with a 20m trajectory and a 10m interval. The resulting submaps are downsampled using 
a voxel grid with grid size of 0.2m. In total 20,731 point clouds are collected for training. 
As the provided ground truth poses are not accurate enough to obtain cross-sequence point-to-point 
correspondences, we generate training samples with synthetic transformations: for a given point 
cloud we create another one by applying an arbitrary rotation around the upright axis and then 
adding Gaussian noise \(\mathcal{N}(0,\sigma_{noise})\) with \(\sigma_{noise}=0.02m\). Note that as 
a point cloud is centered wrt. its centroid before entering the network, no translation is added to 
the synthetic transformations. 
For the global part, we use the dataset proposed in PointNetVLAD~\cite{pointnetvlad}. Specifically, 
for each of the 23 full traversals out of the 44 selected sequences from Oxford RobotCar, a testing 
reference map is generated consisting of the submaps extracted in the testing part of the 
trajectory at 20m intervals. More details on preparing the training data are left to the 
supplementary document.

%\noindent
{\bf Runtime.} For a point cloud with 8192 points, our local (including feature description and 
keypoint detection) and global descriptors can be extracted in one forward pass in 80ms.
As comparison, 3DFeatNet takes 400ms (detection)+510ms (NMS)+18ms (512 local descriptors); 
3DSmoothNet needs 270ms (preprocessing)+144ms (512 local descriptors).

\subsection{3D Keypoint Repeatability}
We use \textit{relative repeatability} to quantify the performance of our keypoint detector. Given 
two point clouds \(\{\mat{P}, {\mat{P}'}\}\) related by a transformation \textrm{T}, a keypoint 
detector detects keypoints \(K = \lbrack K_1, K_2, \cdots, K_m\rbrack \) and \({K'} = \lbrack 
{K}_1', {K}_2', \cdots, {K}_m'\rbrack \) from them. \(K_i \in K\) is repeatable if the distance 
between \(\textrm{T}({K}_i)\) and its nearest neighbor \({K}_j' \in {K}'\) is less than 0.5m. 
\textit{Relative repeatability} is then defined as \(|K_{rep}| / |K|\) with \(K_{rep}\) the 
repeatable keypoints. We use the Oxford RobotCar testing set provided by 3DFeatNet~\cite{3dfeat}, 
which contains 3426 point cloud pairs constructed from 794 point clouds. We compare to three 
handcrafted 3D detectors, ISS~\cite{iss3d}, SIFT-3D~\cite{sift3d} and 
Harris-3D~\cite{sipiran2011harris} and two learned ones 3DFeatNet~\cite{3dfeat} and 
USIP~\cite{li2019usip}. The results are presented in Fig.~\ref{fig:repeat}. As the most recently 
proposed learning based 3D detector that is dedicatedly designed for feature repeatability, USIP 
apparently dominates this benchmark. It is worth noting that the keypoints detected by USIP are 
highly clustered, which is partially in favor of achieving a high repeatability. Moreover, USIP is 
a pure detector, while 3DFeatNet and ours learn detection and description at the same time. 
Our detector outperforms all the other competitors by a large margin when detecting more than 64 
keypoints. In the case of 256 keypoints, our repeatability is roughly 1.75x than the best follower 
3DFeatNet. This clearly demonstrates that, when learning detector and descriptors together, 
\textit{describe-and-detect} is superior than \textit{detect-then-describe}. It is yet interesting 
to see how the key ideas of USIP~\cite{li2019usip} can be merged into this concept.

\subsection{Point Cloud Registration}\label{subsec:reg}
\begin{figure}[t]
	\begin{floatrow}
		\ffigbox{%
	\centering
	\includegraphics[width=0.5\textwidth]{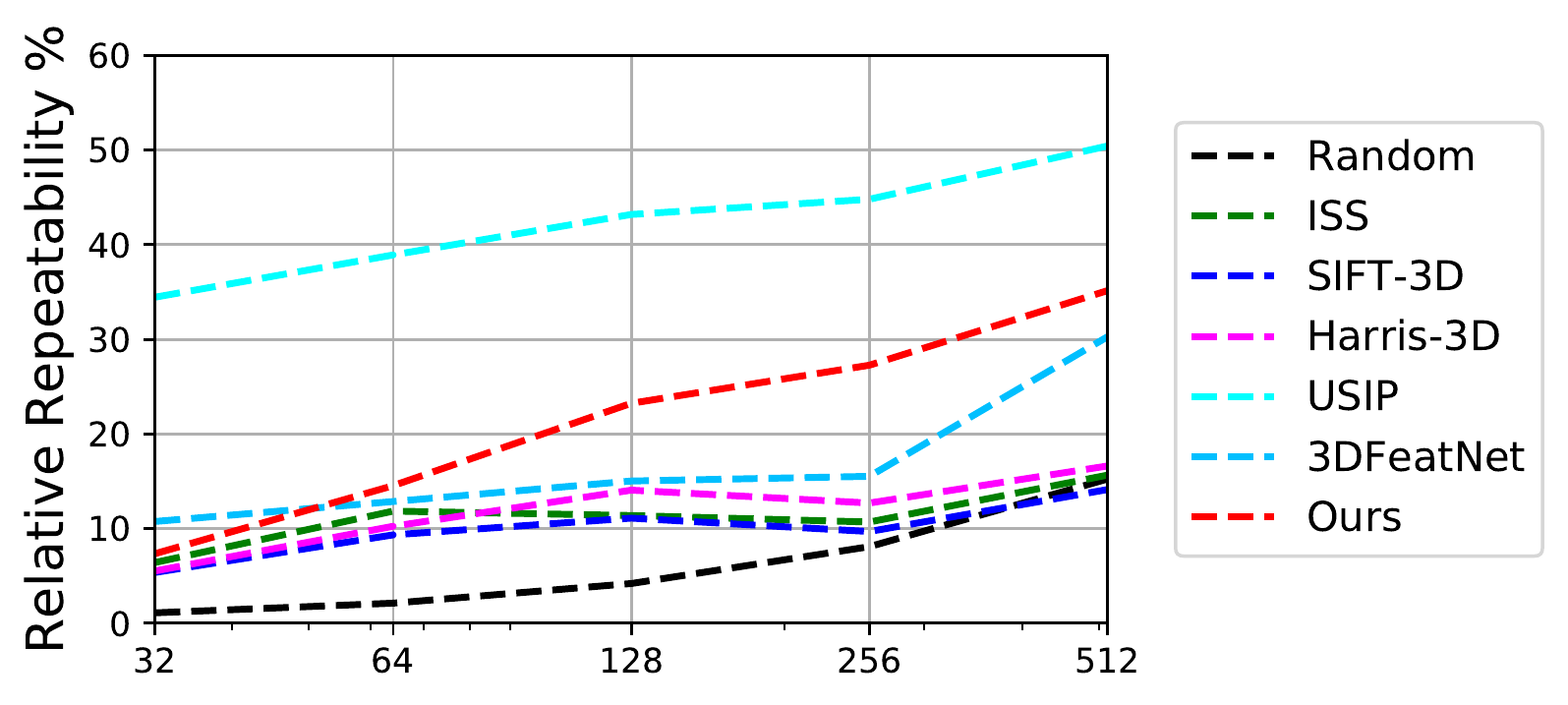}
        }{%
	\caption{Relative repeatability when different number of keypoints are detected.}
	\label{fig:repeat}
		}
		\ffigbox{%
			\centering
			\includegraphics[width=1.\linewidth]{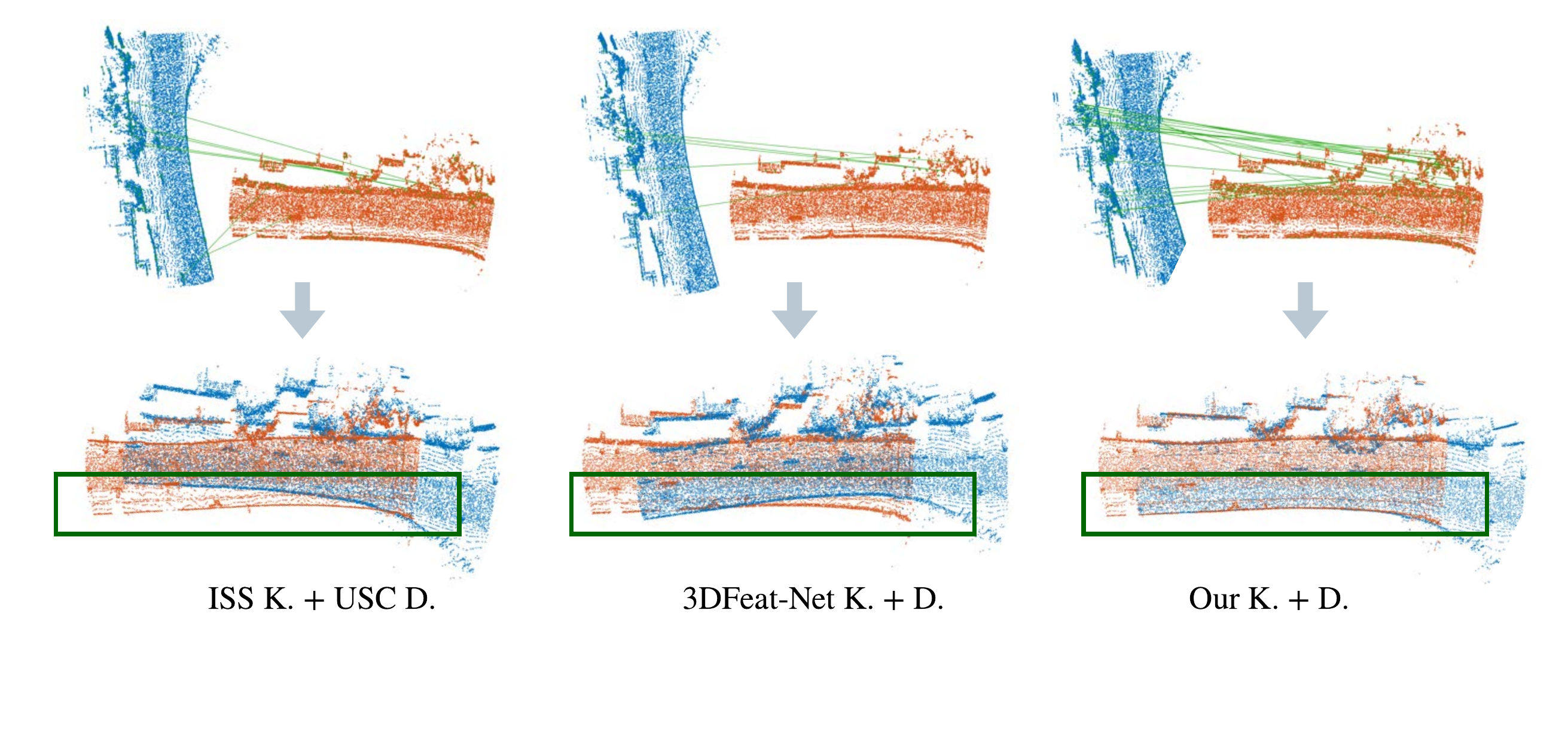}
		}{%
			\caption{Qualitative point cloud registration on Oxford RobotCar. Green lines show the 
			inliers of RANSAC.}
			\label{fig:reg}
		}
	\end{floatrow}
\end{figure}

\begin{table}[t]
	\scalebox{.7}{
		\centering
		\caption{Point cloud registration performance on Oxford RobotCar. Each row in the table 
		corresponds to a keypoint detector and each column refers to a local 3D descriptor. In each 
		cell, we show Relative Translational Error (RTE), Relative Rotation Error (RRE), the 
		registration success rate and the average number of RANSAC iterations. The methods are 
		evaluated on the testing set provided by~\cite{3dfeat}. The top three results of each 
		metric are highlighted in \cbbest{best}/\cbsecond{2nd best}/\cbthird{3rd 
		best}.}
		\label{tab:local_robotcar}
		\begin{tabular}{l|c|c|c|c|c}
			\toprule
			& \multicolumn{5}{c}{RTE (m) / RRE($\si{\degree}$) / Succ. (\%) / Iter.} \\
			\toprule
			& FPFH & 3DSmoothNet & 3DFeatNet & FCGF & DH3D \\
			\midrule
			Random & 
            0.44/1.84/89.8/7135 & 
			0.34/1.39/96.2/7274 & 
			0.43/1.62/90.5/9898 & 
			0.61/ 2.01/39.87/7737 
			& 0.33/1.31/92.1/6873 \\
			ISS & 
			0.39/1.60/92.3/7171 & 
			0.32/1.21/96.8/6301 & 
			0.31/1.08/97.7/7127 & 
			0.56/1.89/43.99/7799 
			& \cbthird{0.30}/1.04/97.9/4986 \\
			Harris-3D & 
			0.54/2.31/47.5/9997 & 
			0.31/1.19/97.4/5236 & 
			0.35/1.33/95.0/9214 & 
			0.57/1.99/46.82/7636 
			& 0.34/1.20/96.4/5985 \\
			3DFeatNet & 
			0.43/2.01/73.7/9603 & 
			0.34/1.34/95.1/7280 & 
			\cbthird{0.30}/1.07/98.1/2940 & 
			0.55/1.89/43.35/5958 & 
			0.32/1.24/95.4/2489 \\
			USIP & 
			0.36/1.55/84.3/5663 & 
			\cbsecond{0.28}/\cbsecond{0.93}/98.0/\cbsecond{584} & 
			\cbsecond{0.28}/\cbbest{0.81}/\cbbest{99.1}/\cbbest{523} & 
			0.41/1.73/53.42/3678 & 
			\cbthird{0.30}/1.21/96.5/\cbthird{1537} \\
			\midrule
			DH3D & 
			0.75/1.85/55.6/8697 & 
			0.32/1.22/96.0/3904 & 
			\cbsecond{0.28}/1.04/\cbthird{98.2}/2908 & 
			0.38/1.48/49.47/4069 & 
			\cbbest{0.23}/\cbthird{0.95}/\cbsecond{98.5}/1972 \\
			\bottomrule
		\end{tabular}	
	}
\end{table}

\begin{table}[t]
	\scalebox{.7}{
		\centering
		\caption{Point cloud registration performance on ETH. The top three 
		results of each metric are highlighted in \cbbest{best}/\cbsecond{2nd best}/\cbthird{3rd 
		best}. Note that RANSAC does not converge within the max. iterations (10000) with 
		FCGF.}
		\begin{tabular}{l|c|c|c|c|c}
			\toprule
			& \multicolumn{5}{c}{RTE (m) / RRE($\si{\degree}$) / Succ. (\%) / Iter.} \\
			\toprule
			& SI & 3DSmoothNet & 3DFeatNet & FCGF & DH3D \\
			\midrule
			Random & 
			0.36/4.36/95.2/7535 & 
			\cbthird{0.18}/2.73/\cbbest{100}/\cbsecond{986} & 
			0.30/4.06/95.2/6898 & 
			0.69/52.87/17.46/10000 & 
			0.25/3.47/\cbbest{100}/5685 \\
			ISS & 
			0.37/5.07/93.7/7706 &  
			\cbbest{0.15}/\cbthird{2.40}/\cbbest{100}/\cbsecond{986} & 
			0.31/3.86/90.5/6518 & 
			0.65/24.78/6.35/10000 & 
			0.19/2.80/93.8/3635 \\
			Harris-3D & 
			0.35/4.83/90.5/8122 & 
			\cbbest{0.15}/2.41/\cbbest{100}/\cbbest{788} & 
			0.27/3.96/88.9/6472 & 
			0.43/55.70/6.35/10000 & 
			0.22/3.47/93.4/4524 \\
			3DFeatNet & 
			0.35/5.77/87.3/7424 & 
			\cbsecond{0.17}/2.73/\cbbest{100}/1795 & 
			0.33/4.50/95.2/6058 & 
			0.52/47.02/3.17/10000 & 
			0.27/3.58/93.7/6462 \\
			USIP & 
			0.32/4.06/92.1/6900/ & 
			\cbthird{0.18}/2.61/\cbbest{100}/\cbthird{1604} & 
			0.31/3.49/82.5/7060 & 
			0.54/27.62/15.87/10000 & 
			0.29/3.29/95.2/4312 \\
			\midrule
			DH3D & 
			0.42/4.65/81.3/7922 & 
			0.38/3.49/\cbbest{100}/5108 & 
			0.36/\cbsecond{2.38}/\cbthird{95.5}/3421 & 
			0.56/48.01/15.87/10000 & 
			0.3/\cbbest{2.02}/\cbsecond{95.7}/3107 \\
			\bottomrule
		\end{tabular}
		\label{tab:local_eth}
	}
\end{table}

Geometric registration is used to evaluate 3D feature matching. A SE3 transformation is estimated 
based on the matched keypoints using RANSAC. We compute Relative 
Translational Error (RTE) and Relative Rotation Error (RRE) and consider a registration is 
successful when RTE and RRE are below 2m and \(5^\circ\), respectively. We compare to two 
handcrafted (ISS~\cite{iss3d} and Harris-3D~\cite{sipiran2011harris}) and two learned 
(3DFeatNet~\cite{3dfeat} and USIP~\cite{li2019usip}) detectors, and three handcrafted 
(SI~\cite{johnson1997spin}, USC~\cite{tombari2010unique} and FPFH~\cite{rusu2010fast}) and three 
learned (3DSmoothNet~\cite{gojcic2019perfect}, 3DFeatNet~\cite{3dfeat} and 
FCGF~\cite{choy2019fully}) descriptors.
The average RTE, RRE, the success rate and the average number of RANSAC iterations on Oxford 
RobotCar of each detector-descriptor combination are shown in Tab.~\ref{tab:local_robotcar}. Note 
that due to space limitation, only the best performing handcrafted descriptor is shown (the same 
applies to Tab.~\ref{tab:local_eth} and ~\ref{tab:dsores}). Although USIP shows 
significantly better performance on repeatability, our method 
now delivers competitive or even better results when applied for registration, where 
both keypoint detector and local feature encoder are needed. This on one hand demonstrates the 
strong discriminative power of our local descriptors, on the other hand also supports our idea of 
learning detector and descriptors in the \textit{describe-and-detect} manner. Another thing to 
point out is that FCGF was trained on KITTI, which might explain its relatively bad results in this 
evaluation. Some qualitative results can be found in Fig.~\ref{fig:reg}.

Unlike Oxford RobotCar, the ETH dataset~\cite{pomerleau2012challenging} contains largely 
unstructured vegetations and much denser points, therefore is used to test the generalizability. 
The same detectors and descriptors as above are tested and the results 
are shown in Tab.~\ref{tab:local_eth}.
We notice that 3DSmoothNet shows the best performances 
on ETH. One important reason is that 3DSmoothNet adopts a voxel grid of size 0.02m to downsample 
the point clouds, while our DH3D and other methods use size 0.1m. Thus 3DSmoothNet has finer 
resolution and is more likely to produce smaller errors. Apart from that, our detector performs 
fairly well (last row) and when combine with our descriptor, it achieves the smallest rotation 
error. With all the detectors, FCGF descriptors cannot make RANSAC converge within the maximum 
number of iterations. The bad performance of FCGF is also noticed by the authors 
of~\cite{bai2020d3feat} and 
was discussed on their GitHub page\footnote{\url{https://github.com/XuyangBai/D3Feat/issues/1}}.

\subsection{Point Cloud Retrieval} \label{subsec:global}
\begin{figure}
	\begin{floatrow}
		\capbtabbox{%
			\scalebox{.8}{
			\begin{tabular}{p{0.20\textwidth}p{0.12\textwidth}p{0.12\textwidth}}
				\toprule
				Method     & @1\% & @1\\
				\midrule
				PN\_MAX    &  73.44  & 58.46
				\\
				PN\_VLAD   &  81.01  & 62.18
				\\
				PCAN	   &  83.81	 & 69.76
				\\
				\tb{Ours-4096} &\tb{84.26}	&\tb{73.28}
				\\
				\tb{Ours-8192} &\tb{85.30}  &\tb{74.16}
				\\
				\bottomrule
		\end{tabular}
		}
		}{%
			\caption{Average recall (\%) at top 1\% and top 1 for Oxford RobotCar.}%
			\label{tab:global}
		}
		\ffigbox{%
			\centering
			\includegraphics[width=0.7\linewidth]{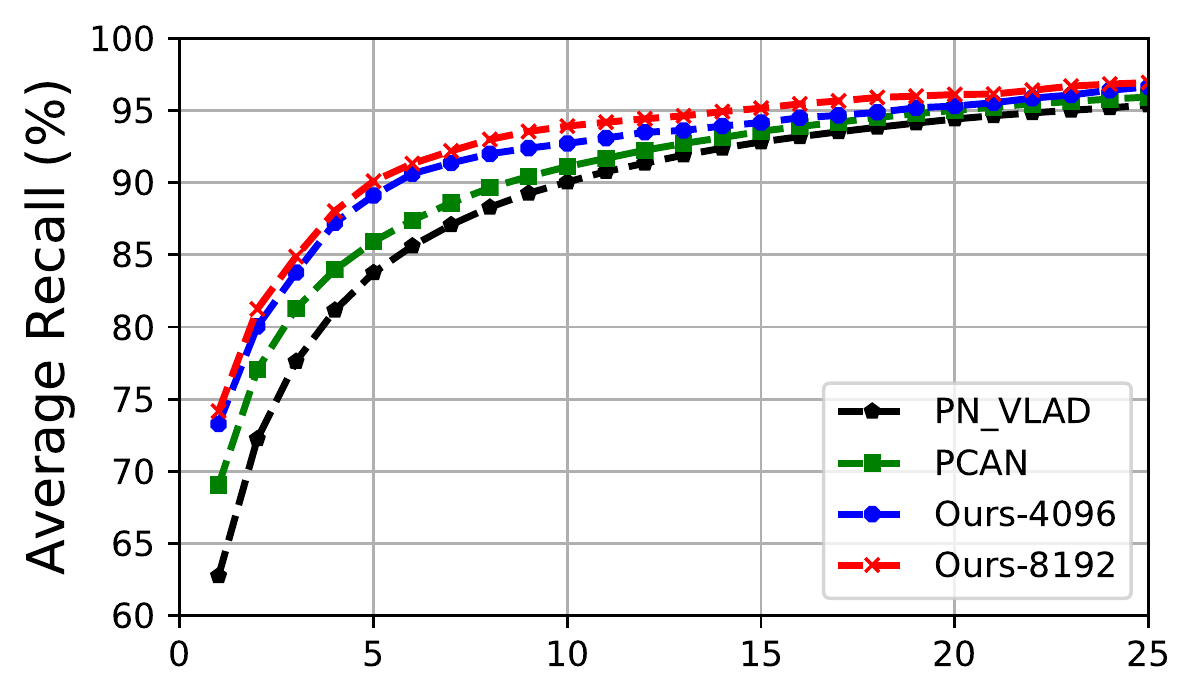}
			}{%
				\caption{Average recall of the top 25 retrievals on Oxford RobotCar.}
				\label{fig:globalretrieval}
			}
	\end{floatrow}
\end{figure}
\noindent
We compare our method against the two state-of-the-art 
approaches, PCAN~\cite{pcan} and PointNetVLAD (PN\_VLAD)~\cite{pointnetvlad}. We also report the 
results of PN\_MAX presented in PN\_VLAD, which consists of the original PointNet architecture with 
the maxpool layer and a fully connected layer to produce a global descriptor. Note that both 
PN\_VLAD and PCAN take submaps of size 4096, whereas ours takes 8192 to favor local feature 
extraction. We thus add a downsampling layer before the final NetVLAD layer to make 
sure the same size of 4096 points enter the final aggregation procedure. For demonstration, we also 
report the results of using the default setting.
We first evaluate the average recall at top 1\% and top1 and show the results in 
Tab.~\ref{tab:global}. Our method with both settings outperforms all the other methods. We further 
show the recall curves of the top25 retrieval matches in Fig.~\ref{fig:globalretrieval}, where 
both of our networks consistently outperform the other two state-of-the-art approaches. The 
evaluation results prove that our network can effectively take advantage of the informative local 
features and produce more discriminative global descriptors. 
Also note that even with the number of points halved before entering the NetVLAD layer, the 
performance drop of our method is very small, which shows that our local features integrate 
sufficient contextual information for global retrieval. Some qualitative retrieval results are 
provided in the supplementary document.

\subsection{Application to Visual SLAM}
\label{app_SLAM}
\begin{table}[htbp]
	\scalebox{.7}{
		\centering
		\caption{Generalization of point cloud registration for visual SLAM. In this experiment, 
		point clouds are generated by running Stereo DSO\cite{stereodso} on Oxford. For learning 
		based methods, models trained on LiDAR points are used without fine-tuning. The top three 
		results of each metric are highlighted in 
		\cbbest{best}/\cbsecond{2nd best}/\cbthird{3rd best}.}
		\begin{tabular}{l|c|c|c|c|c}
			\toprule
			& \multicolumn{5}{c}{RTE (m) / RRE($\si{\degree}$) / Succ. (\%) / Iter.} \\
			\toprule
			& FPFH & 3DSmoothNet & 3DFeatNet & FCGF & DH3D \\
			\midrule
			Random & 
			0.56/2.82/53.13/9030 & 
			0.70/2.19/73.1/6109 & 
			0.72/2.37/69.0/9661 & 
			0.51/2.65/74.93/5613 & 
			0.70/2.23/71.9/7565 \\
			ISS & 
			0.56/3.03/43.58/9210 & 
			0.67/2.15/79.1/6446 & 
			0.58/2.41/71.9/9776 & 
			0.51/2.57/71.94/6015 & 
			0.48/\cbsecond{1.72}/\cbsecond{90.2}/6312 \\
			Harris-3D & 
			0.49/2.67/45.67/9130 & 
			0.48/2.07/74.9/6251 & 
			0.66/2.26/64.5/9528 & 
			0.48/2.63/74.03/5482 & 
			0.39/2.27/68.1/7860 \\
			3DFeatNet & 
			0.62/3.05/35.52/7704 & 
			\cbthird{0.38}/2.22/66.6/5235 & 
			0.92/1.97/84.1/8071 & 
			0.54/2.64/60.90/\cbsecond{4409} & 
			0.74/2.38/80.9/7124 \\
			USIP & 
			0.54/2.98/48.96/7248 & 
			0.39/2.27/77.3/5593 & 
			0.85/2.24/69.9/8389 & 
			0.51/2.65/67.46/\cbbest{3846} & 
			0.65/2.45/68.1/6824 \\
			\midrule
			DH3D & 
			0.60/2.92/48.96/8914 & 
			\cbbest{0.35}/2.01/77.9/5764 & 
			0.41/\cbthird{1.84}/\cbthird{89.3}/7818 & 
			0.48/2.43/69.55/\cbthird{5002} & 
			\cbsecond{0.36}/\cbbest{1.58}/\cbbest{90.6}/7071 \\		
			\bottomrule
		\end{tabular}
		\label{tab:dsores}
	}
\end{table}

\begin{figure}[htbp]
	\centering
	\includegraphics[width=0.21\textwidth,
	height=1.4cm]{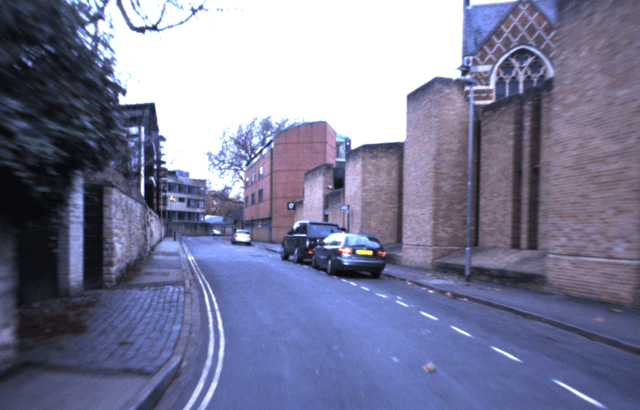}
	\includegraphics[width=0.21\textwidth,
	height=1.4cm]{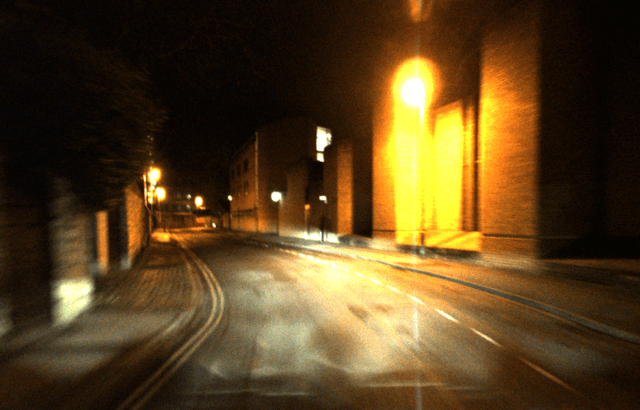}
	\includegraphics[width=0.25\textwidth,
	height=1.4cm]{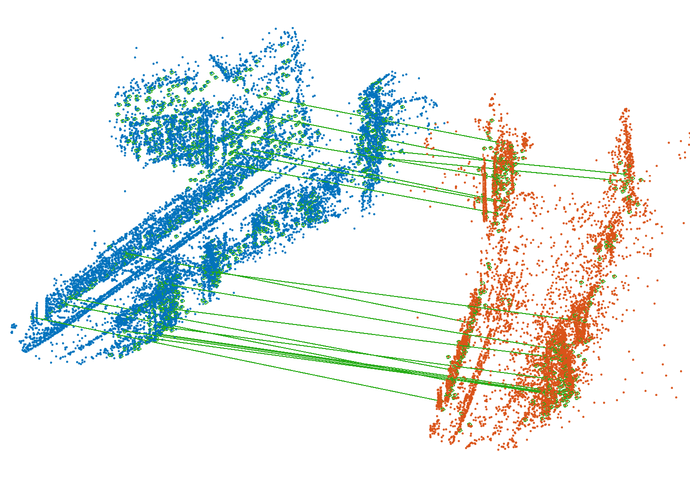}
	\includegraphics[width=0.25\textwidth,
	height=1.4cm]{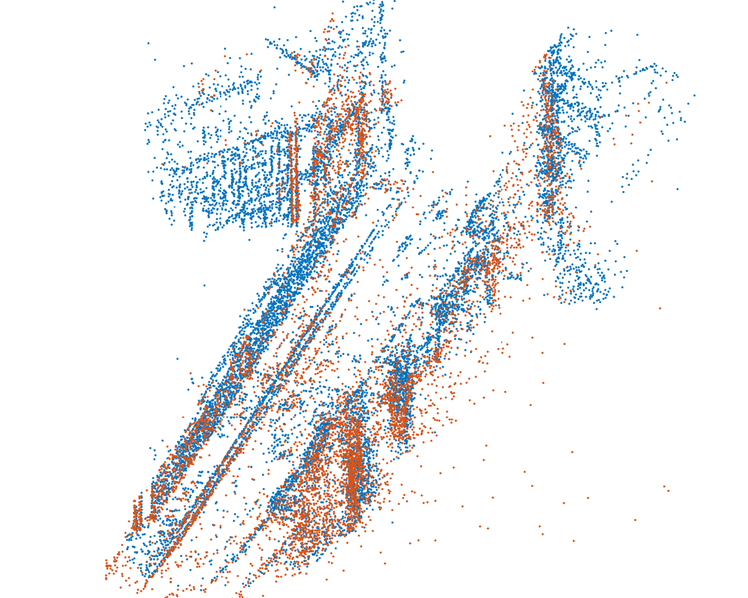}

	\includegraphics[width=0.21\textwidth,
	height=1.4cm]{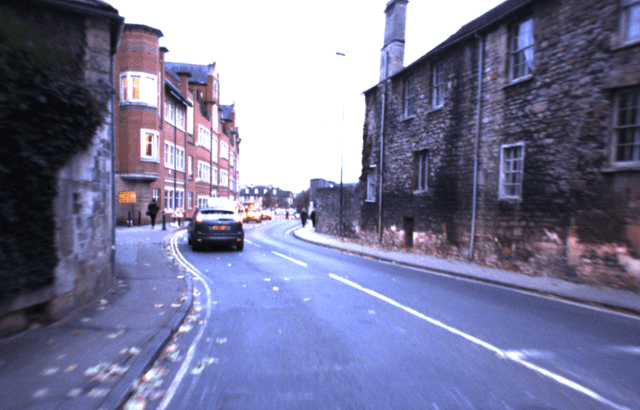}
	\includegraphics[width=0.21\textwidth,
	height=1.4cm]{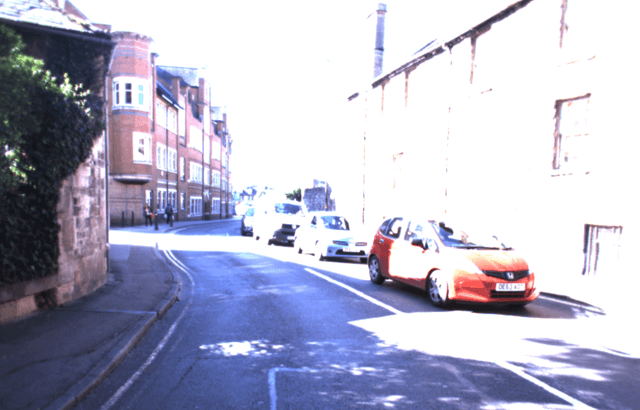}
	\includegraphics[width=0.25\textwidth,
	height=1.4cm]{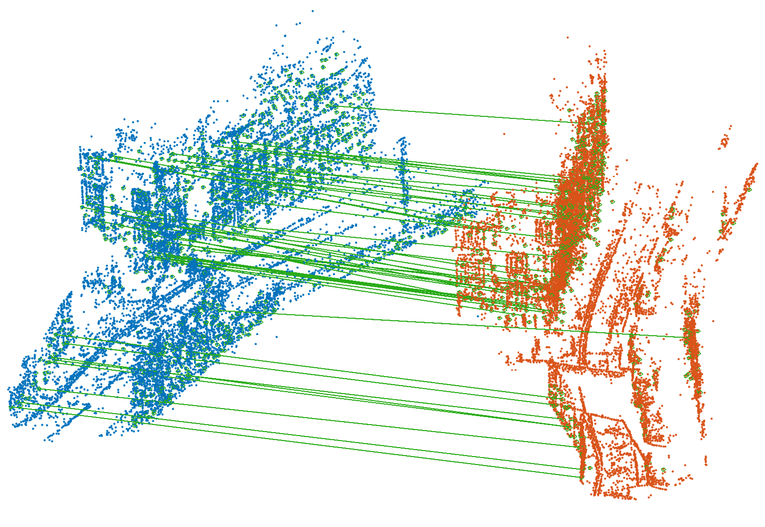}
	\includegraphics[width=0.25\textwidth,
	height=1.4cm]{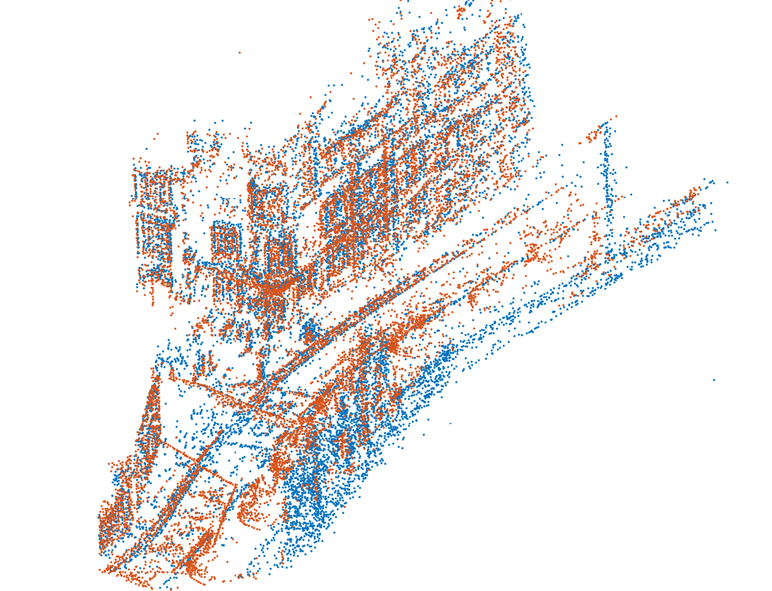}

	\includegraphics[width=0.21\textwidth,
	height=1.4cm]{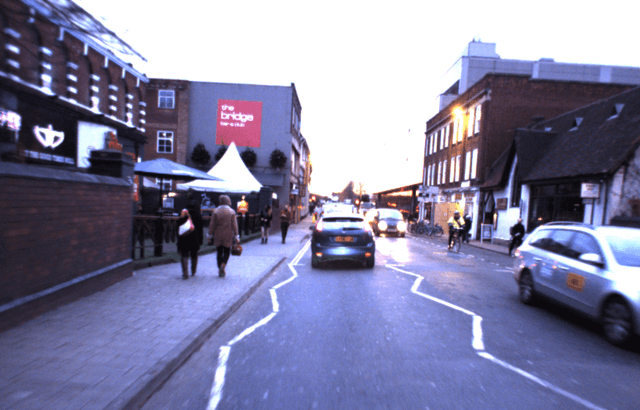}
	\includegraphics[width=0.21\textwidth,
	height=1.4cm]{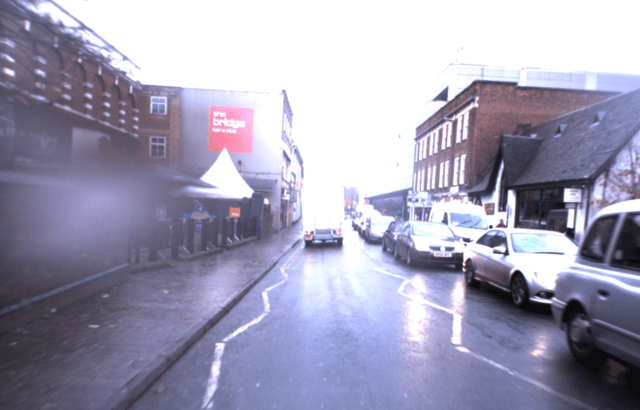}
	\includegraphics[width=0.25\textwidth,
	height=1.4cm]{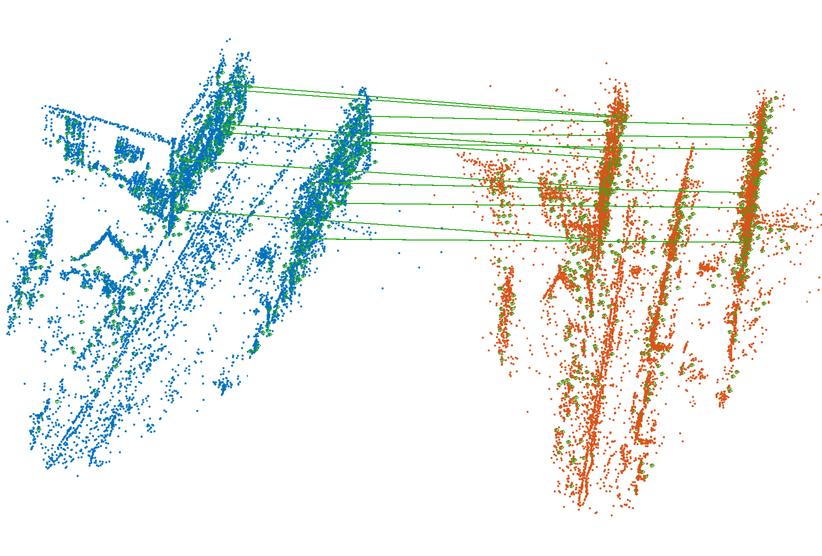}
	\includegraphics[width=0.25\textwidth,
	height=1.4cm]{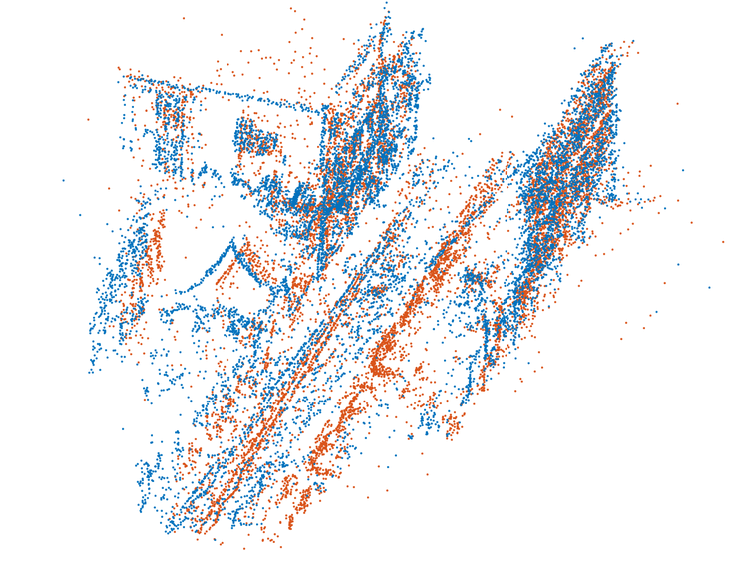}

	\includegraphics[width=0.21\textwidth,
	height=1.4cm]{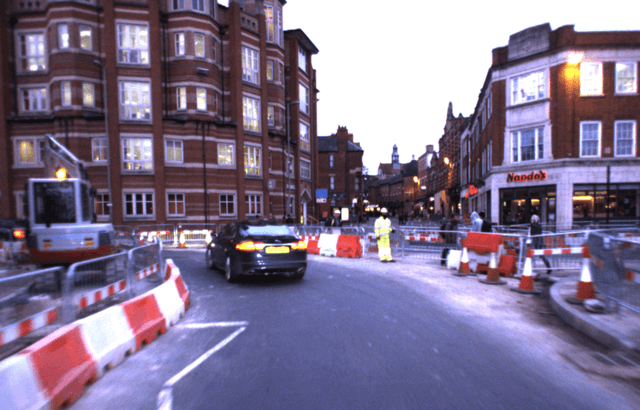}
	\includegraphics[width=0.21\textwidth,
	height=1.4cm]{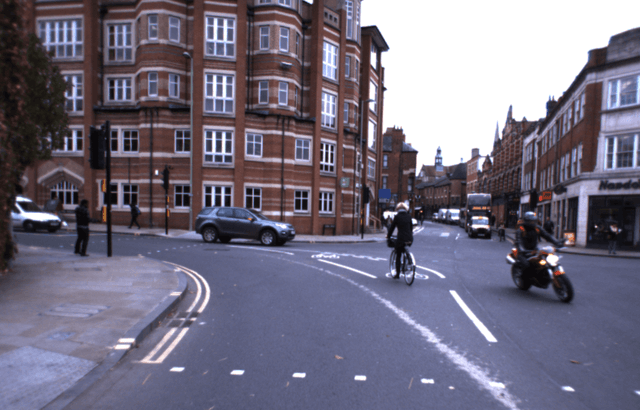}
	\includegraphics[width=0.25\textwidth,
	height=1.4cm]{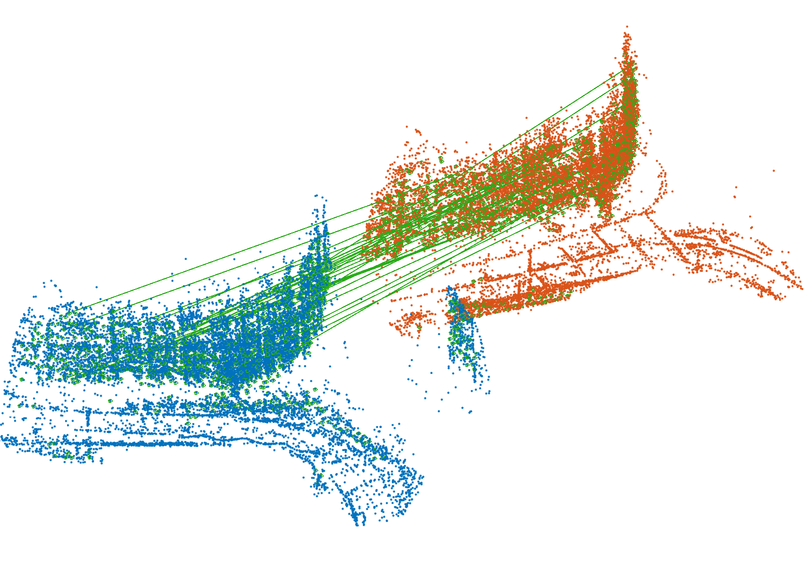}
	\includegraphics[width=0.25\textwidth,
	height=1.4cm]{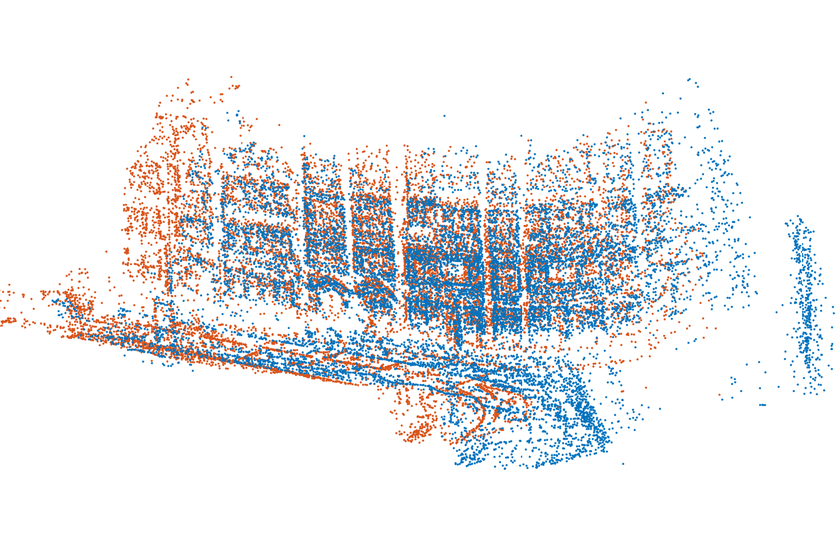}

	\caption{Registration on point clouds generated by Stereo DSO~\cite{stereodso}. The 
	first two columns display frames from the reference and the query sequences. The last two 
	columns show the matches found by RANSAC and the point clouds after alignment.}
	\label{fig:dso_vis}
\end{figure}
\noindent
In this section, we demonstrate the generalization capability of our method to a different sensor 
modality by evaluating its performance on the point clouds generated by Stereo 
DSO~\cite{stereodso}.
As a direct method, Stereo DSO has the advantage of delivering relatively dense 3D 
reconstructions, which can provide stable geometric structures that are less affected by 
image appearance changes. We therefore believe that it is worth exploring extracting 3D descriptors 
from such reconstructions which can be helpful for loop-closure and relocalization for visual SLAM.
To justify this idea, Stereo DSO is used to generate point clouds of eight traversals from Oxford 
RobotCar, which covers a wide range of day-time and weather conditions. This gives us 318 point 
cloud pairs 
with manually annotated relative poses. Each point cloud is cropped with a radius of 30m and 
randomly rotated around the vertical axis. We use the same parameters as in Sec.~\ref{subsec:reg} 
without fine-tuning our network and evaluate the geometric registration performance against other 
methods in Tab.~\ref{tab:dsores}. 
As shown in the table, our approach achieves the best rotation error (1.58$\si{\degree}$) and 
success rate (90.6\%) and second best translation error (0.36m) among all the evaluated methods. It 
can also be noticed that most evaluated methods show significant inferior performances compared to 
the results in Tab.~\ref{tab:local_robotcar}, e.g., the successful rates of 3DFeatNet+3DFeatNet, 
USIP+3DSmoothNet and USIP+3DFeatNet drop from 98.1\%, 98.0\% and 99.1\% to 84.1\%, 77.3\% and 
69.9\%, respectively. This is largely because the point clouds extracted from LiDAR 
scannings have quite different distributions as those from Stereo DSO. Our model is still able to 
achieve a successful rate of 90.6\%, showing the least degree of degeneracy. This further 
demonstrates the good generalization ability of the proposed method. Some qualitative results 
are shown in Fig.~\ref{fig:dso_vis}.

\subsection{Ablation Study}
\begin{figure}[t]
	\begin{floatrow}
		\capbtabbox{%
			\scalebox{.8}{
			\centering
			\begin{tabular}{lcccc}
				\toprule
				Method 			& RTE(m) & RRE($\si{\degree}$)& Succ. & Iter. \\
				\midrule
				w/o \( L_{det}\)& 0.43 & 1.52 & 93.72 & 3713 \\			
				Weak Sup. 		& 0.48 & 1.78 & 90.82 & 3922 \\			
				w/o SE       	& 0.39 & 1.24 & 95.18 & 3628 \\			
				\tb{Default} 	& \tb{0.23} &\tb{0.95} &\tb{98.49} &\tb{1972} \\
				\bottomrule
			\end{tabular}
			}
		}{%
			\caption{Effects of different components for point cloud registration on
				Oxford RobotCar.}%
			\label{tab:components}
		}
		\ffigbox{%
			\centering
			\includegraphics[height=2.3cm]{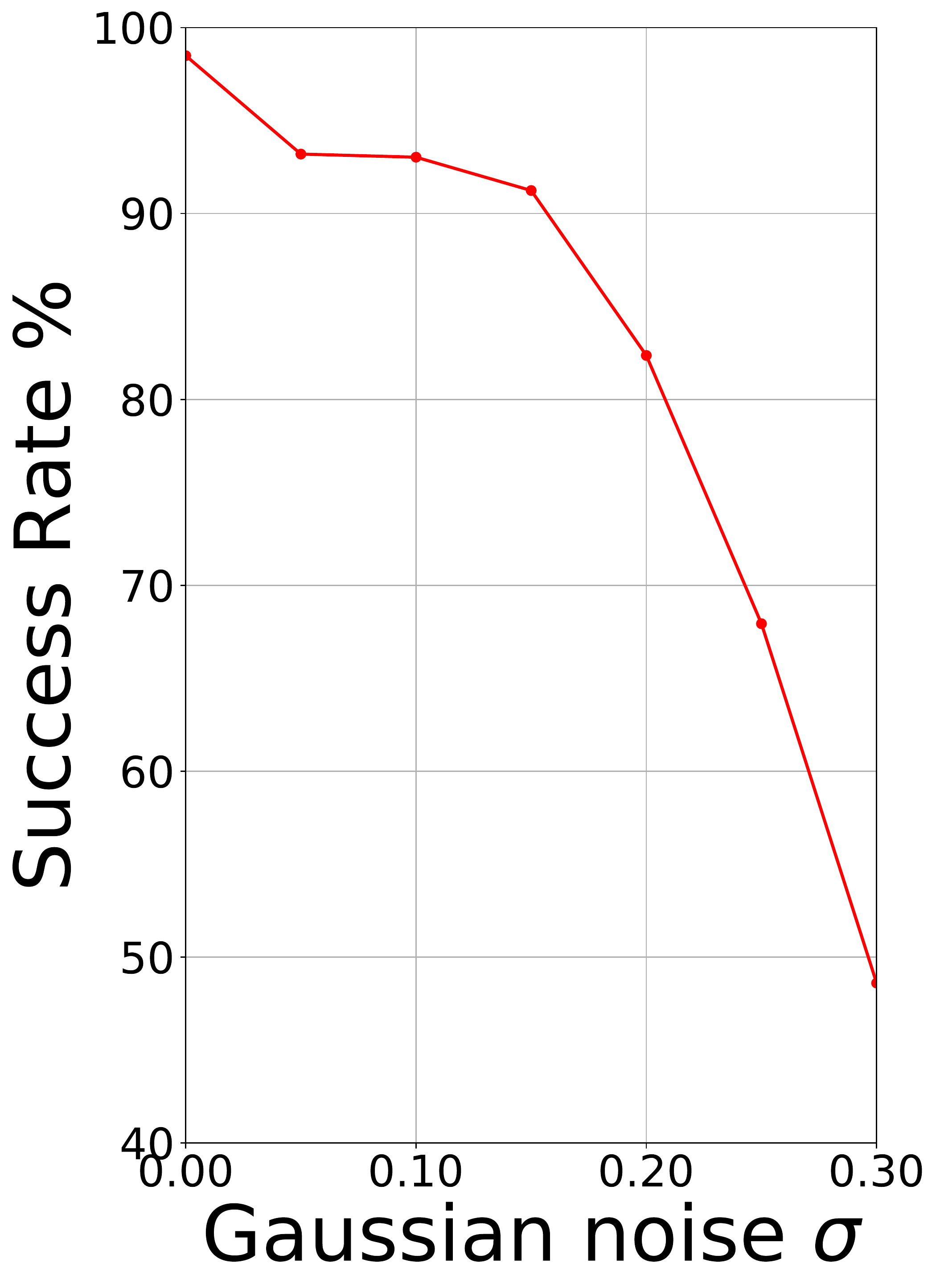}
			\includegraphics[height=2.3cm]{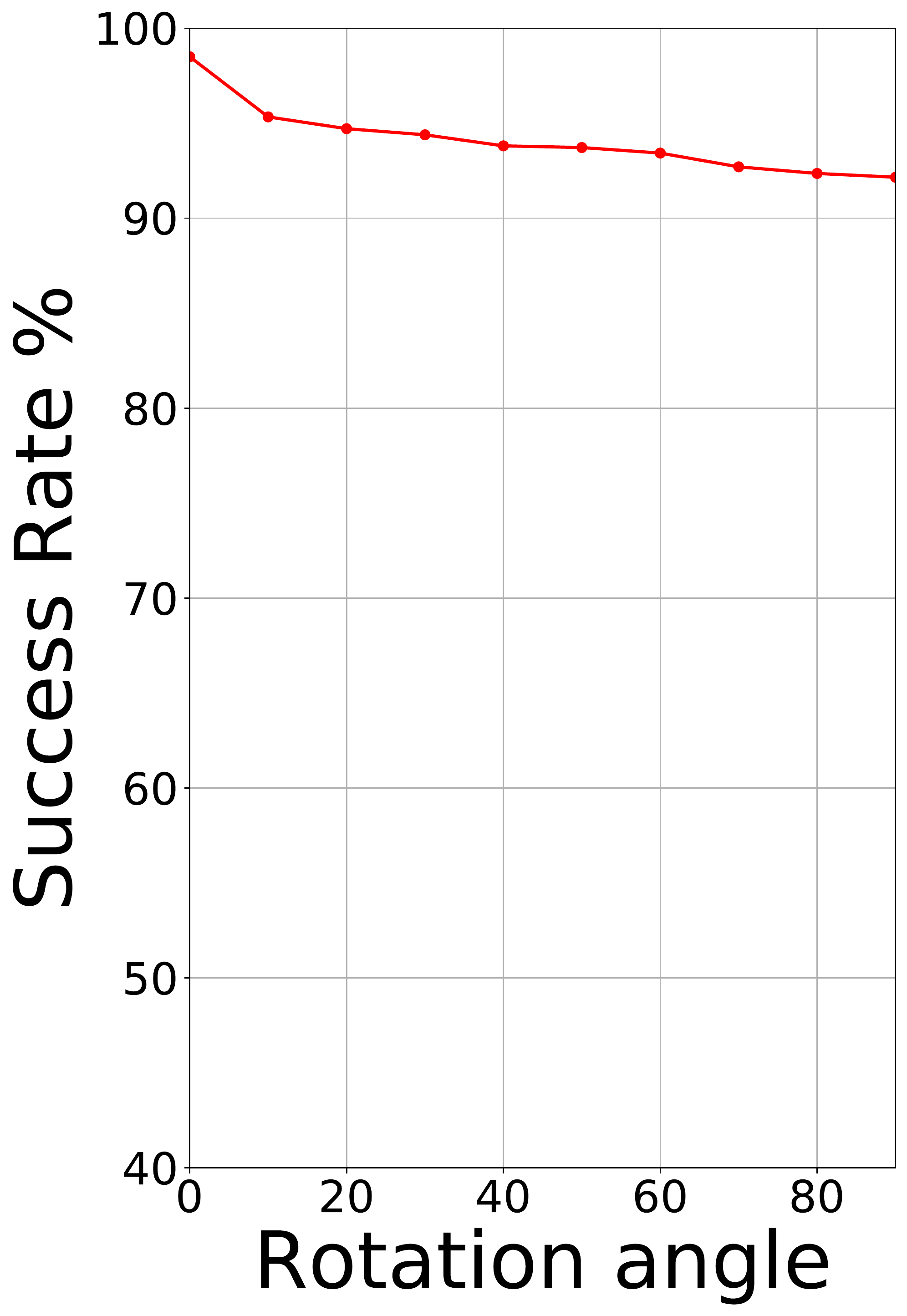}
			\includegraphics[height=2.3cm]{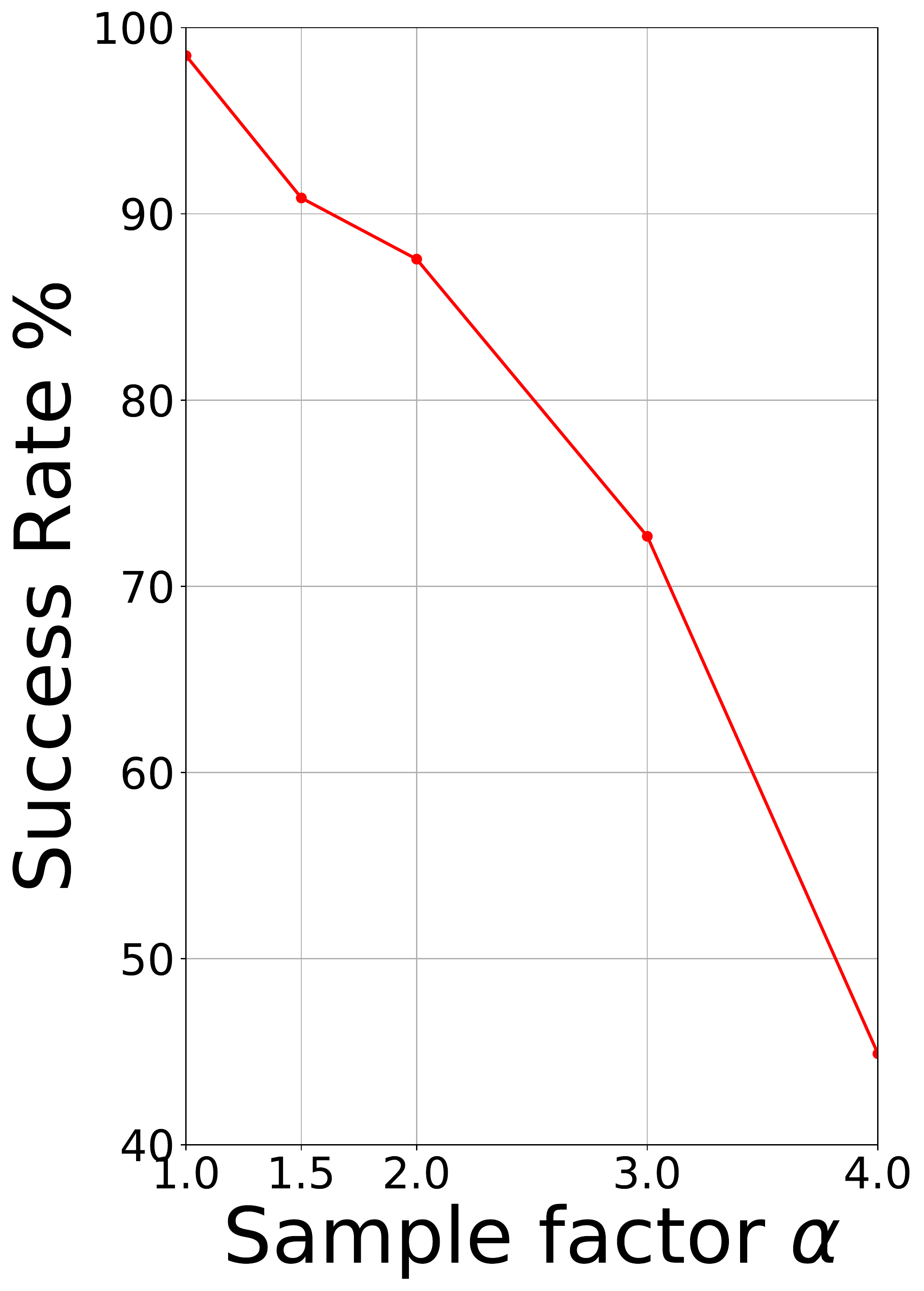}
		}{%
			\caption{Local detector and descriptor robustness test evaluated by the success rate of 
			point cloud registration.}
			\label{fig:ablation_local}
		}
	\end{floatrow}
\end{figure}

\noindent
%-----------------------------------------------------------------------------------------------
{\bf Effectiveness of different components.} We carry out three experiments to explore the 
contributions of different components of our method: (1) We remove the detector module and only 
\(L_{desc}\) is used to train the local feature encoder; (2) The weak supervision at the submap 
level proposed by 3DFeatNet~\cite{3dfeat} is used (details in the supplementary material); (3) We 
remove the SE blocks. As shown in Tab.~\ref{tab:components}, the largest performance 
decrease comes with (2), which verifies our idea of generating a supervision signal by synthesizing 
transformations. Results of (1) indicate that learning an effective 
confidence map \(S\) can improve the quality of the learned local descriptors for matching. The 
results of (3) show that SE blocks contribute to learning more informative local descriptors and 
therefore are helpful to 3D feature matching.

\begin{figure}[t]
	\centering
	\subfloat[Noise]{
		\includegraphics[width=0.23\linewidth]
		{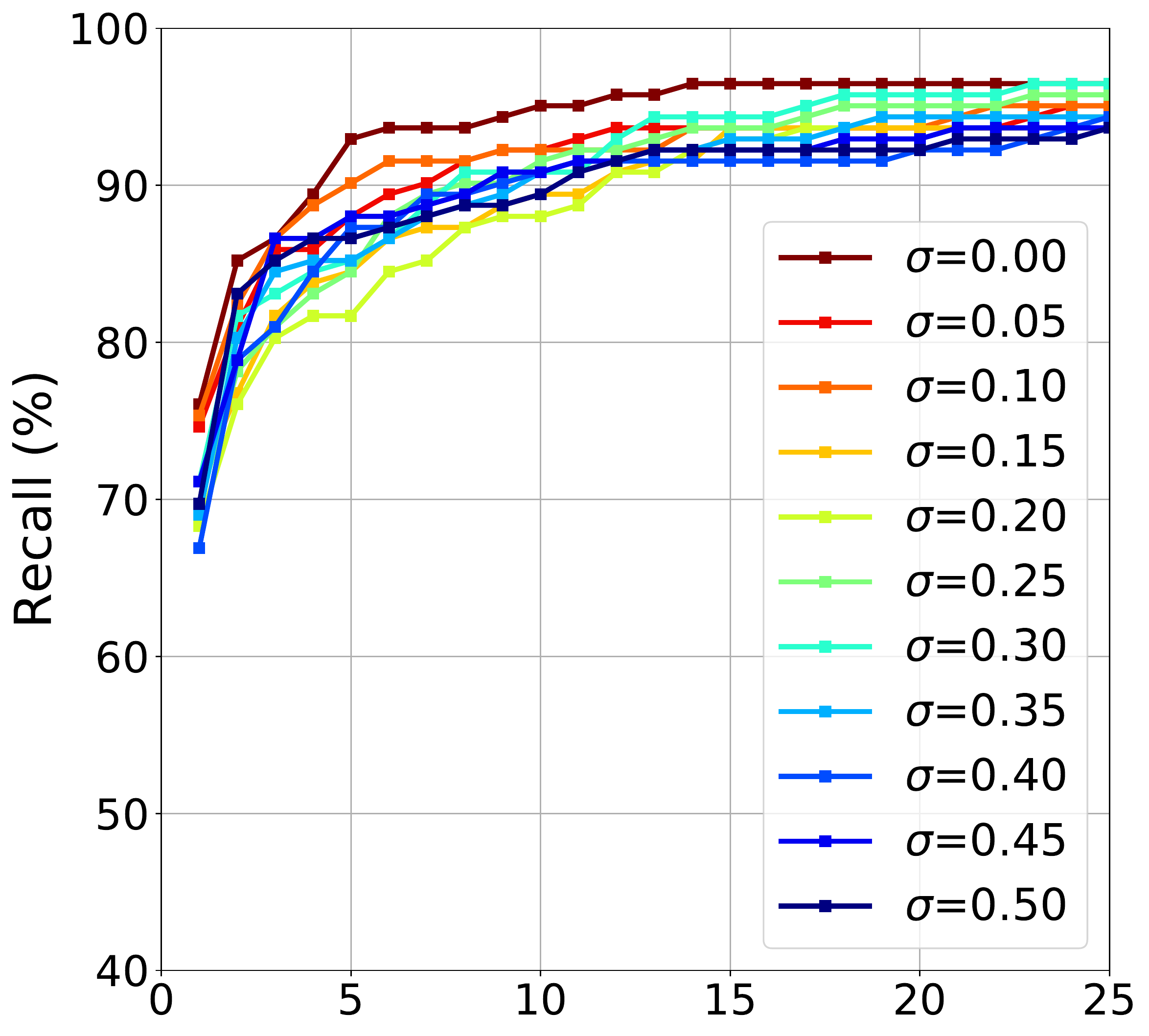}}
	\subfloat[Rotation]{
		\includegraphics[width=0.23\linewidth]
		{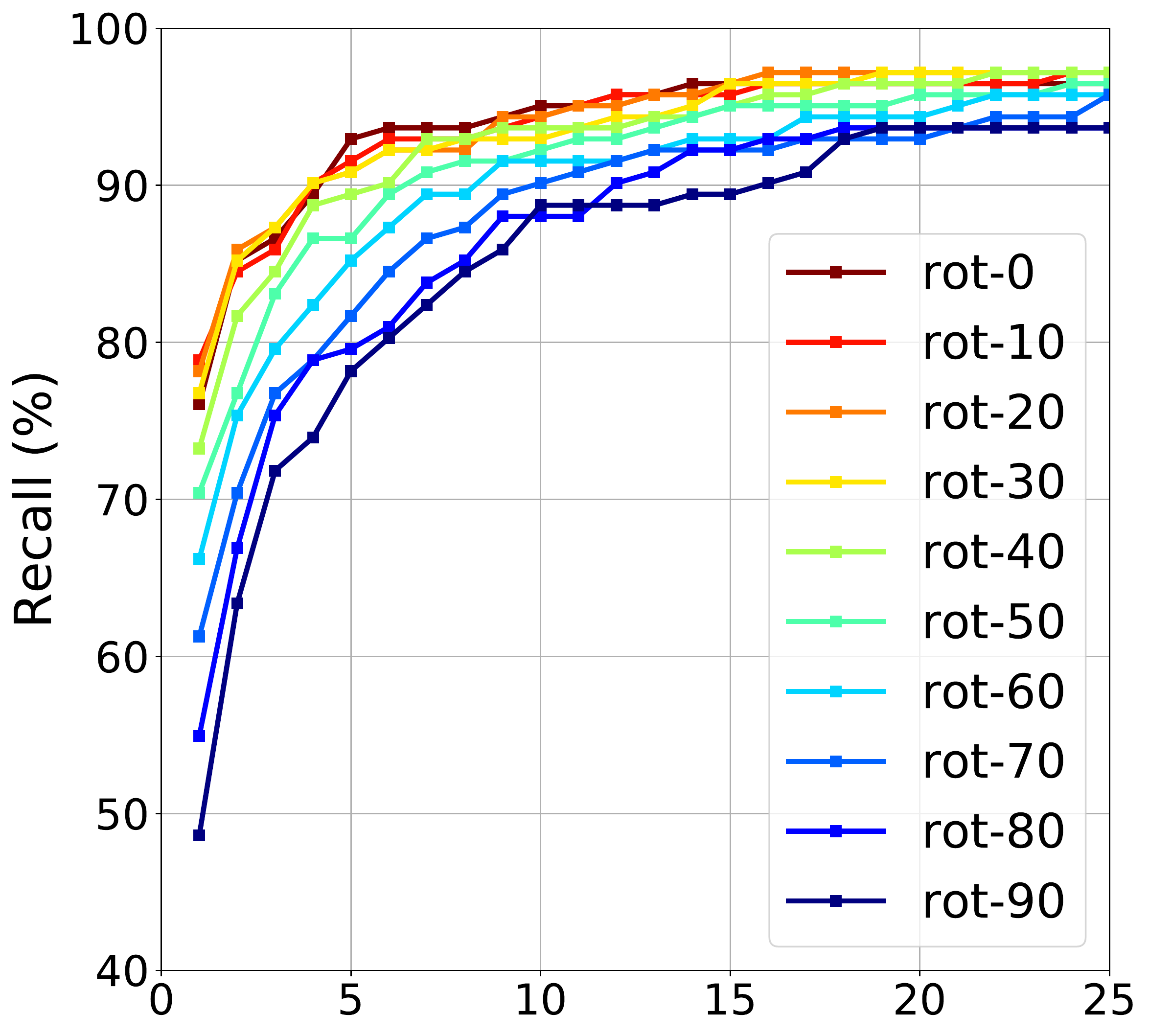}}
	\subfloat[Downsampling]{
		\includegraphics[width=0.23\linewidth]
		{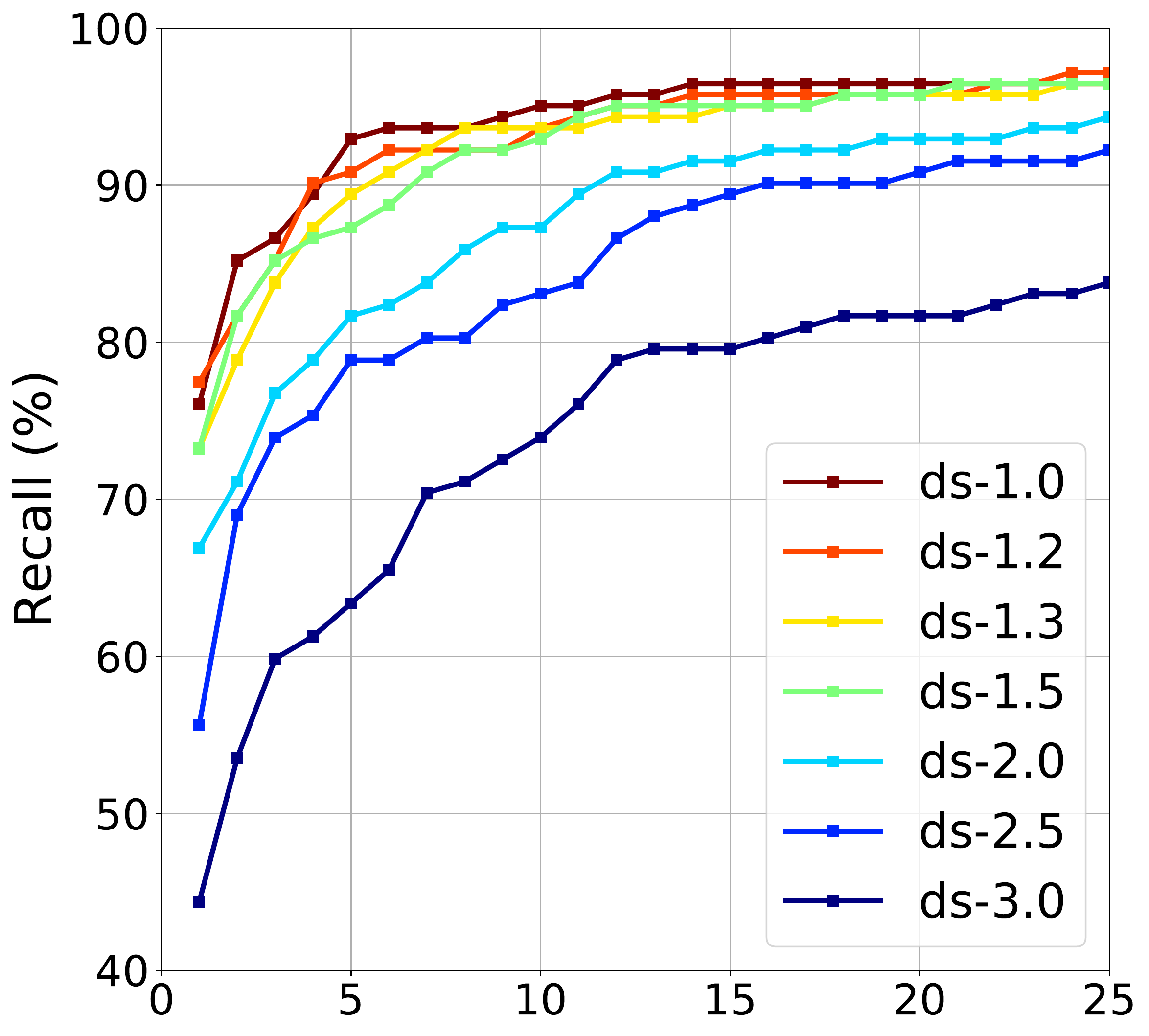}}
	\caption{Global descriptor robustness against random noise, rotation and downsampling. The 
	\(x\) axes show the number of top retrieved matches.}
	\label{fig:ablation_global}
\end{figure}

\medskip
\noindent
%-----------------------------------------------------------------------------------------------
{\bf Robustness test.}
We assess the robustness of our model for both point cloud retrieval and registration against three 
factors, i.e., noise, rotation and downsampling: We add Gaussian 
noise \(\mathcal{N}(0,\sigma_{noise})\) to the point clouds; The range of the rotation test is 
set between 0 and 90\(^\circ\); Point clouds are downsampled using a set of factors \(\alpha\). For 
the local part, as shown in Fig.~\ref{fig:ablation_local}, our descriptors has shown excellent 
rotation invariance. When noise is added, our method can still achieves \(>\) 90\% success rate 
for \(\sigma_{noise} <\) 0.15m. The performance significantly drops for \(\sigma_{noise} >\) 
0.2m, possibly due to the fact that training samples are filtered by a voxel grid with size 0.2m, 
thus strong noise tends to heavily change the underlying point distribution. A similar explanation 
applies to the case of downsampling for a factor \(\alpha > 2\). Nevertheless, our model can 
still guarantee 90\% success rate for \(\alpha \leqslant 1.5\). We conduct the same robustness 
tests for our global descriptors. Fig.~\ref{fig:ablation_global} (a) demonstrates that our global 
descriptors possess good robustness to noise. Contrary to the local descriptors, the global 
descriptor seems to be less robust against rotations, which needs further investigation. Similar to 
the local descriptor, the quality of global feature is not affected too much for \(\alpha 
\leqslant 1.5\).

%%%%%%%%%%%%%%%%%%%%%%%%%%%%%%%%%%%%%%%%%%%%%%%%%%%%%%%%%%%%%%%%%%%%%%%%%%%%%%%%%%%%%%%%%%%%%%%%%%%%
\section{Conclusion}
%%%%%%%%%%%%%%%%%%%%%%%%%%%%%%%%%%%%%%%%%%%%%%%%%%%%%%%%%%%%%%%%%%%%%%%%%%%%%%%%%%%%%%%%%%%%%%%%%%%%
We introduced a hierarchical network for the task of large-scale point cloud based relocalization. 
Rather than pursuing the traditional strategy of detect-then-describe or separately computing local 
and global descriptors, our network performs local feature detection, local feature description and 
global descriptor extraction in a single forward pass. Experimental results demonstrate the
state-of-the-art performance of both our local and global descriptors across multiple benchmarks. 
Our model trained on LiDAR points also shows favorable generalization ability when applied to point 
clouds generated by visual SLAM methods. Future work is focused on further exploring the robustness 
of the learned descriptors to various perturbations.

\clearpage
\bibliographystyle{splncs04}
\bibliography{egbib}
\clearpage

\end{document}

%\renewcommand\thelinenumber{\color[rgb]{0.2,0.5,0.8}\normalfont\sffamily\scriptsize\arabic{linenumber}\color[rgb]{0,0,0}}
% \renewcommand\makeLineNumber {\hss\thelinenumber\ \hspace{6mm} \rlap{\hskip\textwidth\ 
%\hspace{6.5mm}\thelinenumber}}
% \linenumbers
\pagestyle{headings}
\mainmatter
\def\ECCVSubNumber{2754}  % Insert your submission number here

\title{	Supplementary Material\\DH3D: Deep Hierarchical 3D Descriptors for Robust Large-Scale 
	6DoF Relocalization} % Replace with your title

\begin{comment}
\titlerunning{ECCV-20 submission ID \ECCVSubNumber}
\authorrunning{ECCV-20 submission ID \ECCVSubNumber}
\author{Anonymous ECCV submission}
\institute{Paper ID \ECCVSubNumber}
\end{comment}
%******************

%\begin{comment}
\titlerunning{DH3D: Deep Hierarchical 3D Descriptors for 6DoF Relocalization}
\author{Juan Du\inst{1 \star} \and
	Rui Wang\inst{1,2}\thanks{Authors contributed equally.} \and
	Daniel Cremers\inst{1,2}}
\authorrunning{J. Du et al.}
\institute{Technical University of Munich \and
	Artisense\\
	\email{\{duj, wangr, cremers\}@in.tum.de}}
%\end{comment}
%******************
\maketitle	

%%%%%%%%%%%%%%%%%%%%%%%%%%%%%%%%%%%%%%%%%%%%%%%%%%%%%%%%%%%%%%%%%%%%%%%%%%%%%%%%%%%%%%%%%%%%%%%%%%%%
\section{Overview}
%%%%%%%%%%%%%%%%%%%%%%%%%%%%%%%%%%%%%%%%%%%%%%%%%%%%%%%%%%%%%%%%%%%%%%%%%%%%%%%%%%%%%%%%%%%%%%%%%%%%

Despite the significant progresses in deep learning, one issue we are commonly facing is that 
reproducibility is not guaranteed by all the papers. In this respect, in addition to releasing 
our code, we provide in this supplementary document the remaining technical aspects and some 
insights we have gathered during the development. We hope this can help the other researchers in 
this field.

In the first part, we show more detailed results of our keypoint repeatability on the ETH dataset 
and on the point clouds generated by Stereo DSO on Oxford RobotCar (referred to as 
StereoDSO-Oxford), which we omitted in the main paper due to space limitation. We also provide more 
qualitative registration results on Oxford RobotCar and StereoDSO-Oxford, including some failure 
cases. The second part provides results revealing the influence of using different operators for 
global aggregation, where we see that although there exist several ways, attention based NetVLAD 
layer is so far the best choice. Some qualitative results on global retrieval are attached 
afterwards. The third part is dedicated to presenting all the technical details left over from the 
main paper, including (1) structures of sub-networks, (2) training data preparation, (3) two-phase 
training of the local and global networks, (4) generation of the Stereo DSO point clouds, and (5) 
details on the weak supervision used in the ablation study in the main paper.

%%%%%%%%%%%%%%%%%%%%%%%%%%%%%%%%%%%%%%%%%%%%%%%%%%%%%%%%%%%%%%%%%%%%%%%%%%%%%%%%%%%%%%%%%%%%%%%%%%%%
\section{Additional Results on Local Feature}
%%%%%%%%%%%%%%%%%%%%%%%%%%%%%%%%%%%%%%%%%%%%%%%%%%%%%%%%%%%%%%%%%%%%%%%%%%%%%%%%%%%%%%%%%%%%%%%%%%%%

\subsection{Keypoint Repeatability}

\begin{figure}[hbp]
	\centering
	{\includegraphics[width=0.5\textwidth]{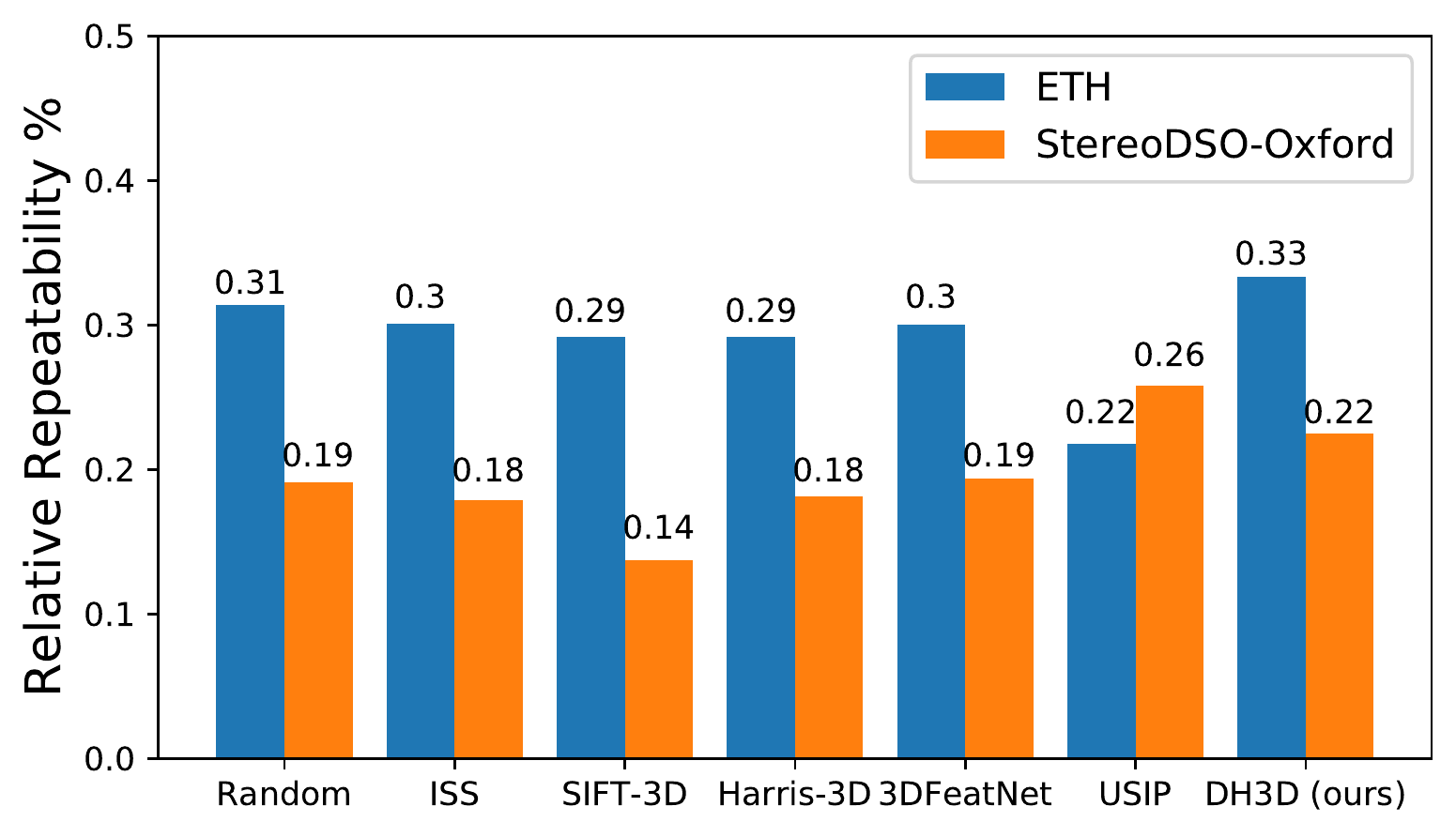}}
	\caption{Relative repeatabily on ETH dataset and StereoDSO-Oxford testing set.}
	\label{fig:rep_eth_dso}
\end{figure}

We use the same method as in the main paper to evaluate keypoint repeatability on ETH and 
StereoDSO-Oxford, except that now we choose 0.3m as the threshold distance for the ETH dataset to 
determine whether a point is repeatable. For all the detectors, 512 and 1024 keypoints are 
respectively extracted for StereoDSO-Oxford and ETH. The results are shown in 
Fig.~\ref{fig:rep_eth_dso}. 

On StereoDSO-Oxford, the results hold a similar pattern as those on the Oxford RobotCar LiDAR 
points shown in the main paper. Our detector outperforms other methods except USIP which is a pure 
detector that dedicatedly designed for feature repeatability. All the handcrafted detectors and 
3DFeatNet have repeatability lower than 0.2. 
On the ETH dataset, our detector has the highest repeatability while the performance of USIP 
degrades significantly. When USIP is trained, a relatively fixed receptive field is defined for 
extracting each feature location (by pre-setting the parameter \textit{M} and \textit{K}, please 
refer to~\cite{li2019usip} for details). In other words, the performance of the learned network is 
highly correlated to certain point densities and scales in the receptive field. Therefore, when 
applied to point clouds with very different spatial distribution as the training data, the network 
suffers to generalize well.

\subsection{Qualitative Results on Oxford RobotCar}
\begin{figure*}[!tbp]
	\centering
	
	\includegraphics[width=0.56\textwidth, 
	height=3.0cm]{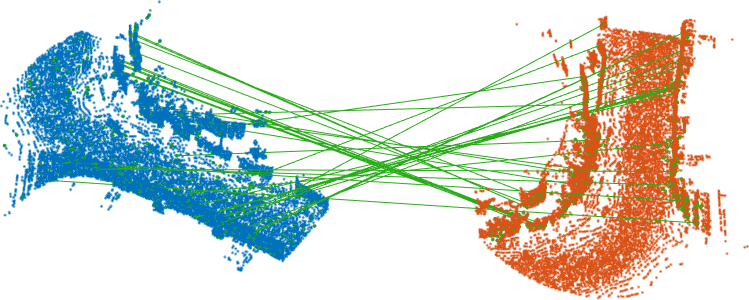}
	\includegraphics[width=0.4\textwidth, 
	height=3.0cm]{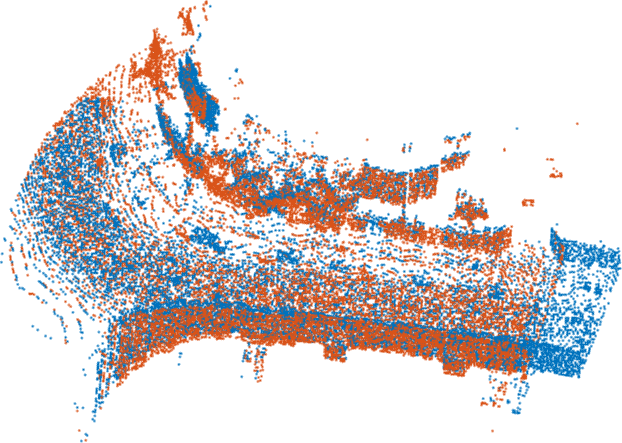}
		\includegraphics[width=0.56\textwidth, 
	height=3.0cm]{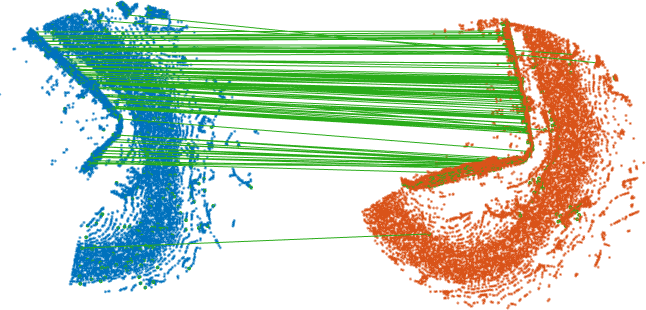}
	\includegraphics[width=0.4\textwidth, 
	height=3.0cm]{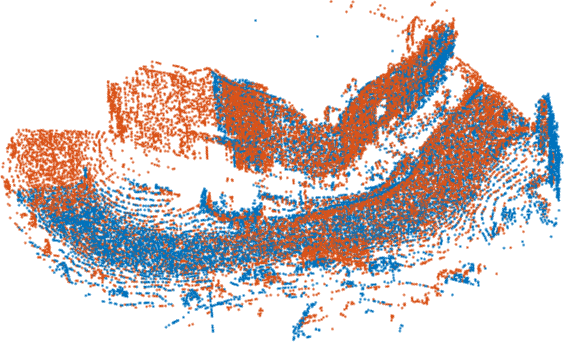}
	\includegraphics[width=0.56\textwidth, 
	height=3.0cm]{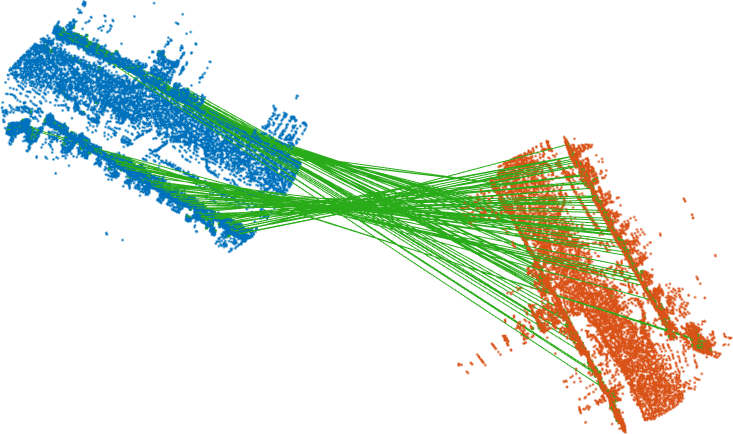}
	\includegraphics[width=0.4\textwidth, 
	height=3.0cm]{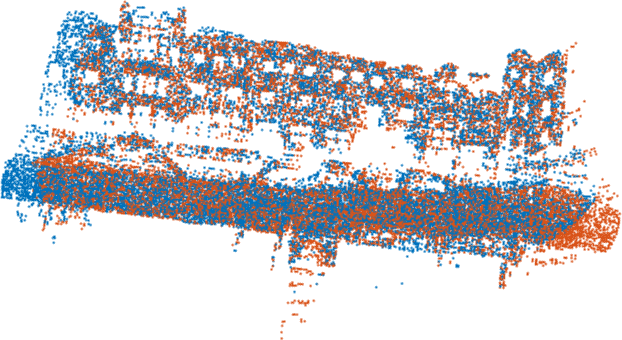}
%	\includegraphics[width=0.56\textwidth, 
%	height=3.0cm]{figures/oxford_res/50_m_cut.png}
%	\includegraphics[width=0.4\textwidth, 
%	height=3.0cm]{figures/oxford_res/50_a_cut.png}
	\includegraphics[width=0.56\textwidth, 
	height=3.0cm]{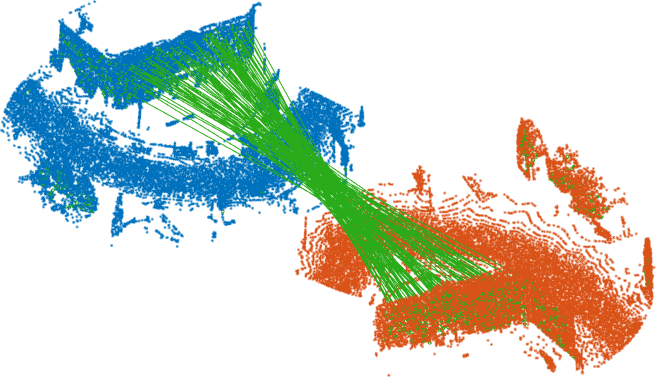}
	\includegraphics[width=0.4\textwidth, 
	height=3.0cm]{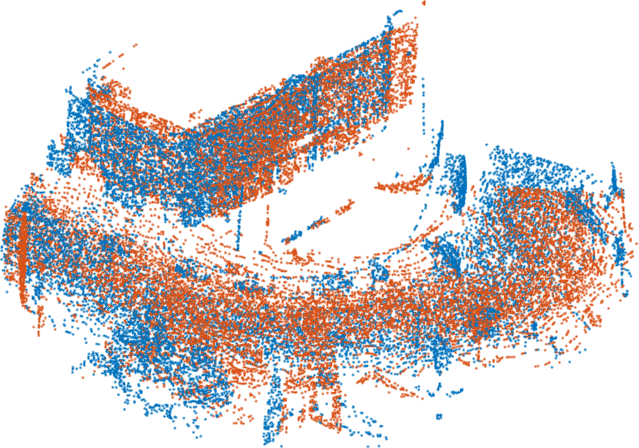}
%	\includegraphics[width=0.24\textwidth, 
%	height=3.0cm]{figures/oxford_res/fail1_g_cut.png}
	\includegraphics[width=0.56\textwidth, 
	height=3.0cm]{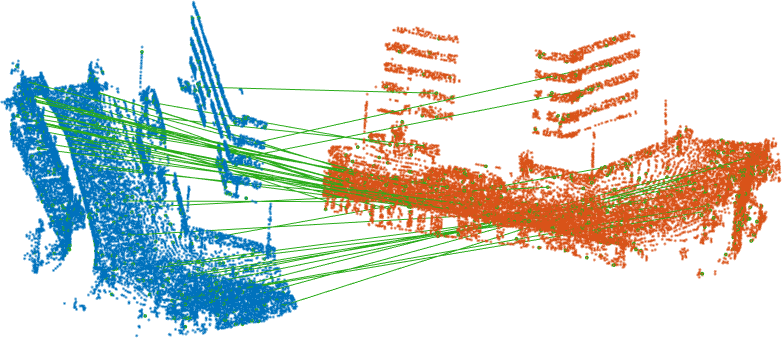}
	\includegraphics[width=0.4\textwidth, 
	height=3.0cm]{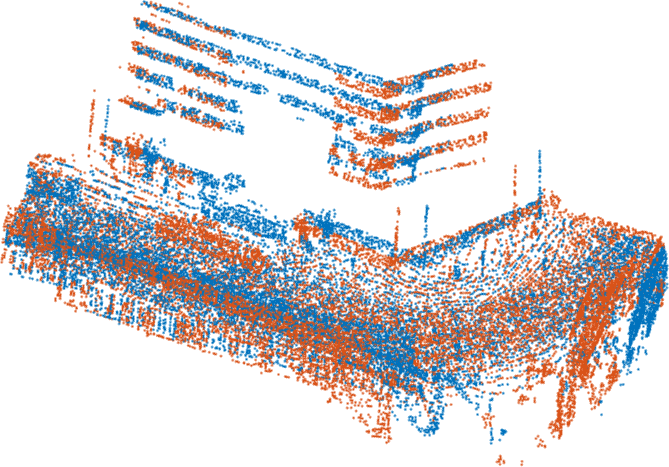}
%	\includegraphics[width=0.24\textwidth, 
%	height=3.0cm]{figures/oxford_res/1890_g_cut.png}
	
	\caption{More results of registration of Oxford RobotCar. The first column shows the 
		matched features (after RANSAC) and the second shows the point clouds after alignment.
		The last two rows show failure cases caused by the dominated repetitive vertical 
		structures.}
	\label{fig:vis_oxford}
\end{figure*}

We show additional qualitative registration results by our method in Fig.~\ref{fig:vis_oxford}, 
including two failure cases due to the domination of local features from the facade of 
buildings which are repeated and contain limited local textures. 

\subsection{Qualitative Results on Stereo DSO Points}
\begin{figure*}[!tbp]
	\centering
	
	\includegraphics[width=0.22\textwidth, 
	height=2.0cm]{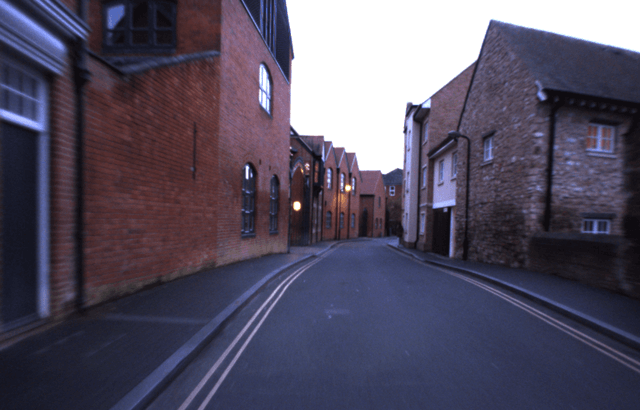}
	\includegraphics[width=0.22\textwidth, 
	height=2.0cm]{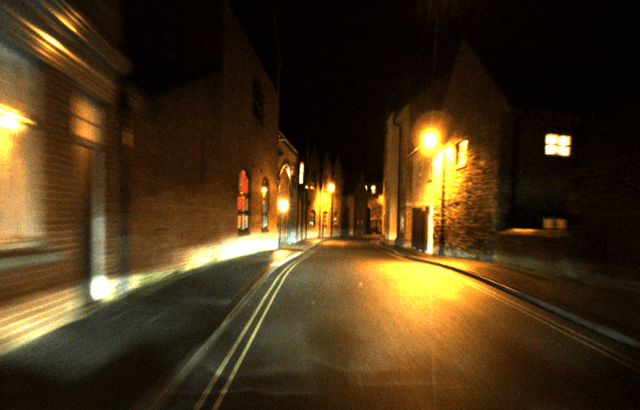}
	\includegraphics[width=0.25\textwidth, 
	height=2.2cm]{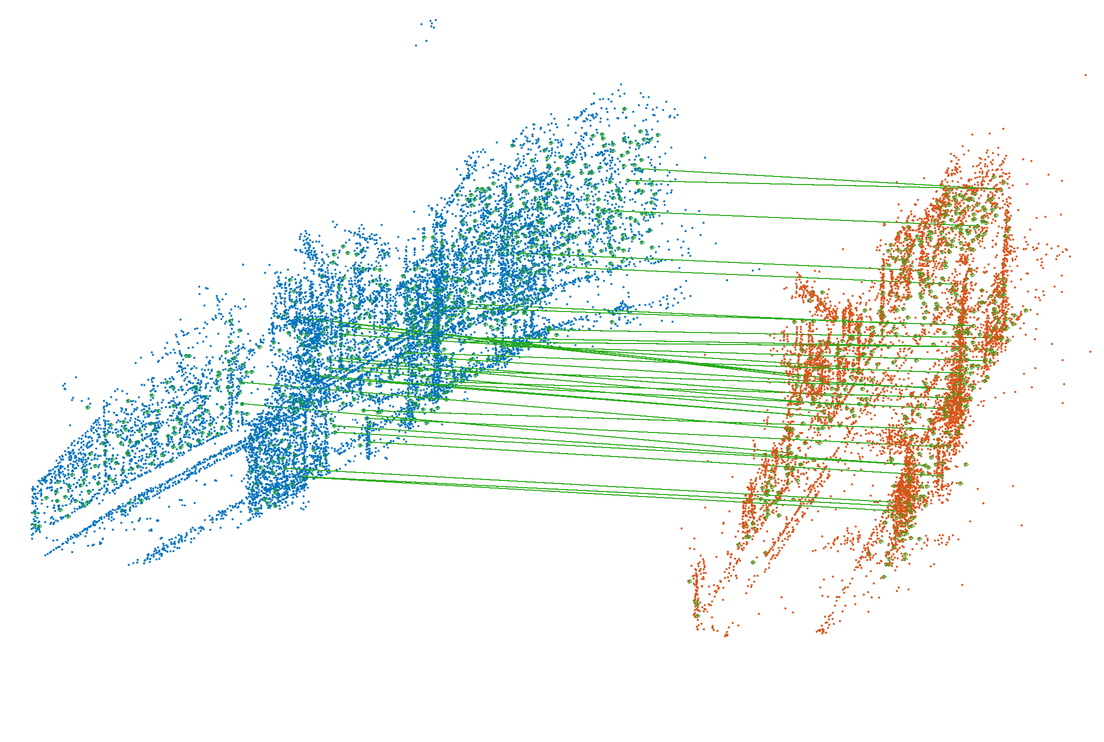}
	\includegraphics[width=0.25\textwidth, 
	height=2.2cm]{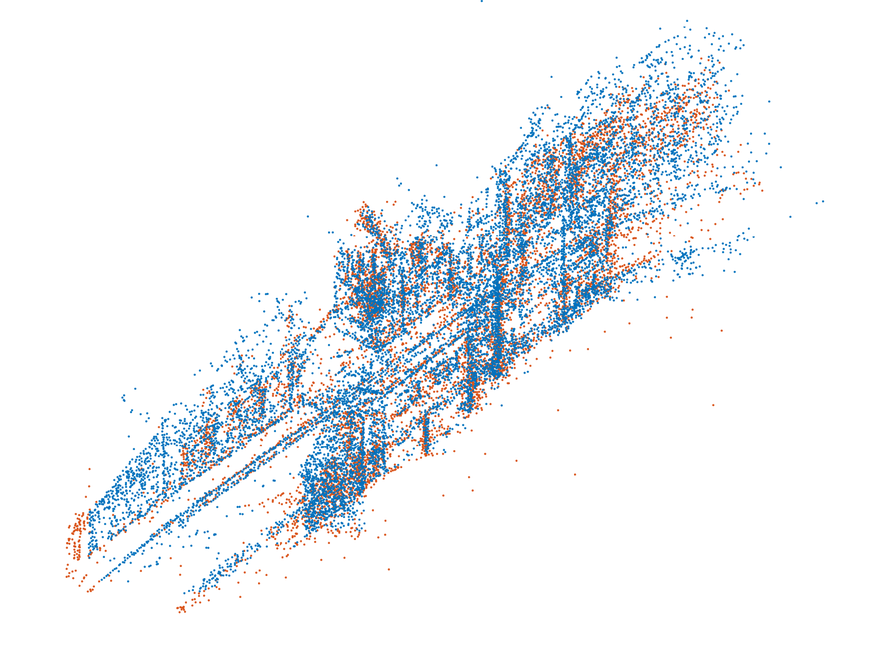}
	
	\includegraphics[width=0.22\textwidth, 
	height=2.0cm]{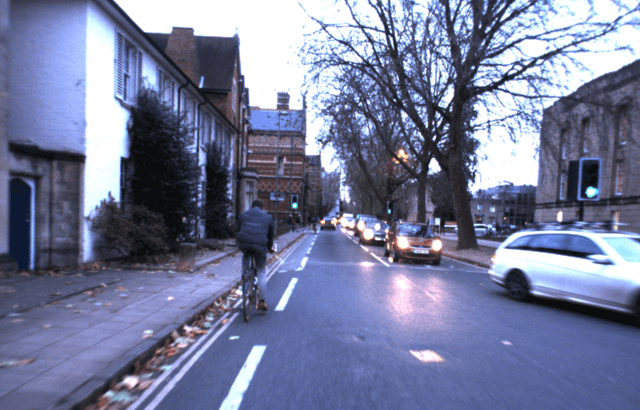}
	\includegraphics[width=0.22\textwidth, 
	height=2.0cm]{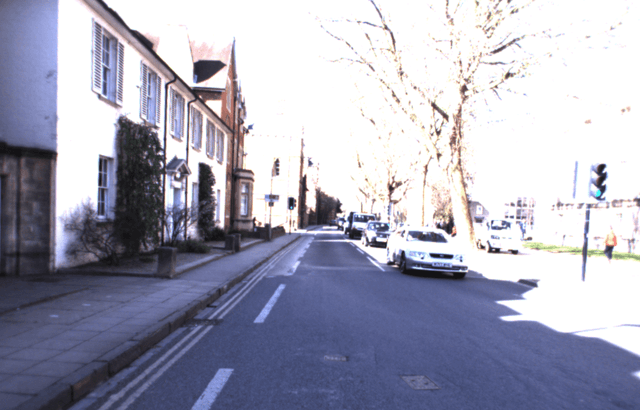}
	\includegraphics[width=0.25\textwidth, 
	height=2.2cm]{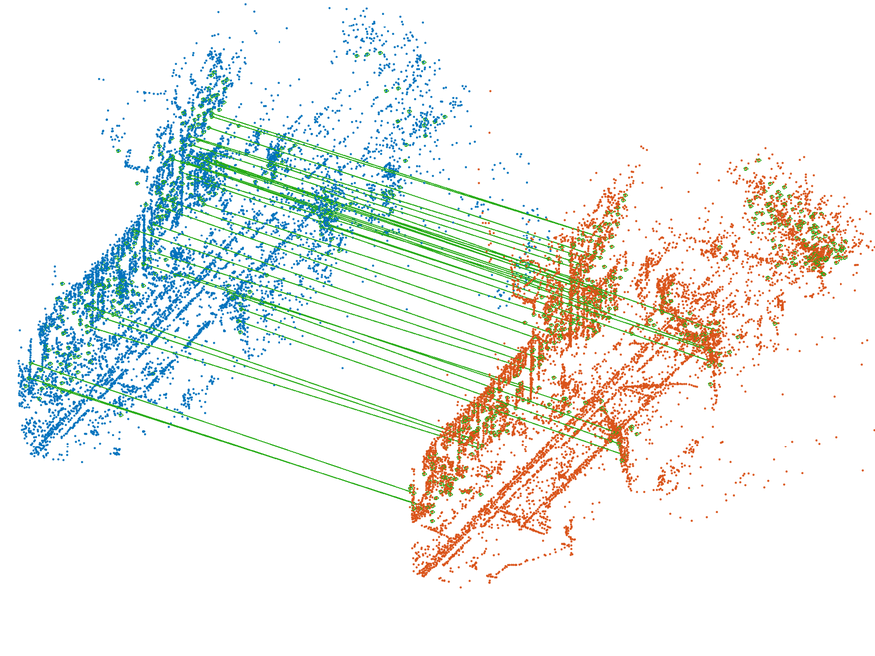}
	\includegraphics[width=0.25\textwidth, 
	height=2.2cm]{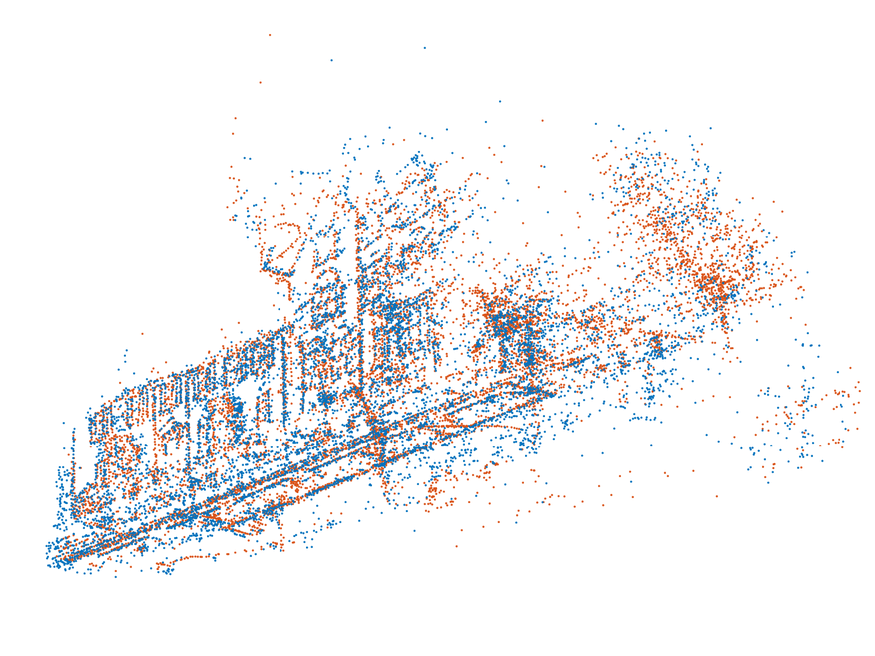}
	
	\includegraphics[width=0.22\textwidth, 
	height=2.0cm]{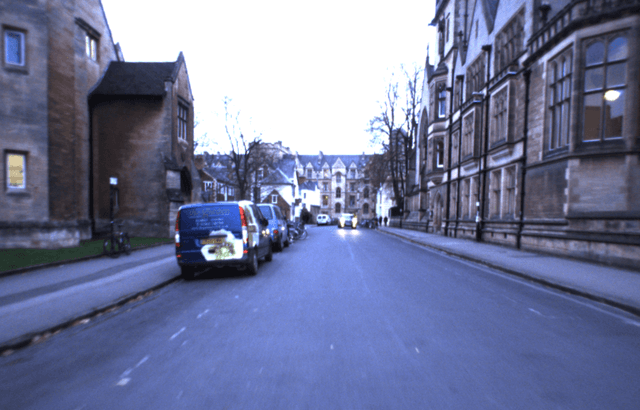}
	\includegraphics[width=0.22\textwidth, 
	height=2.0cm]{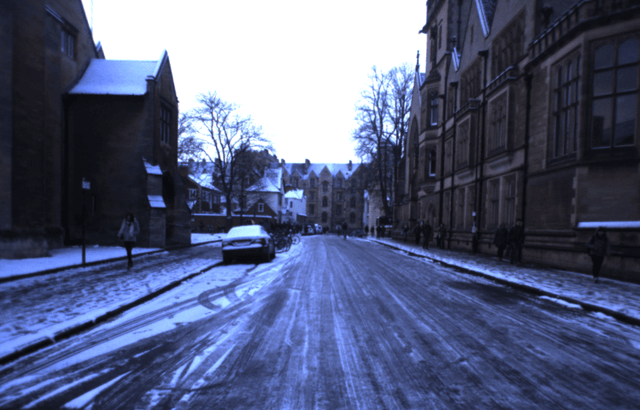}
	\includegraphics[width=0.25\textwidth, 
	height=2.2cm]{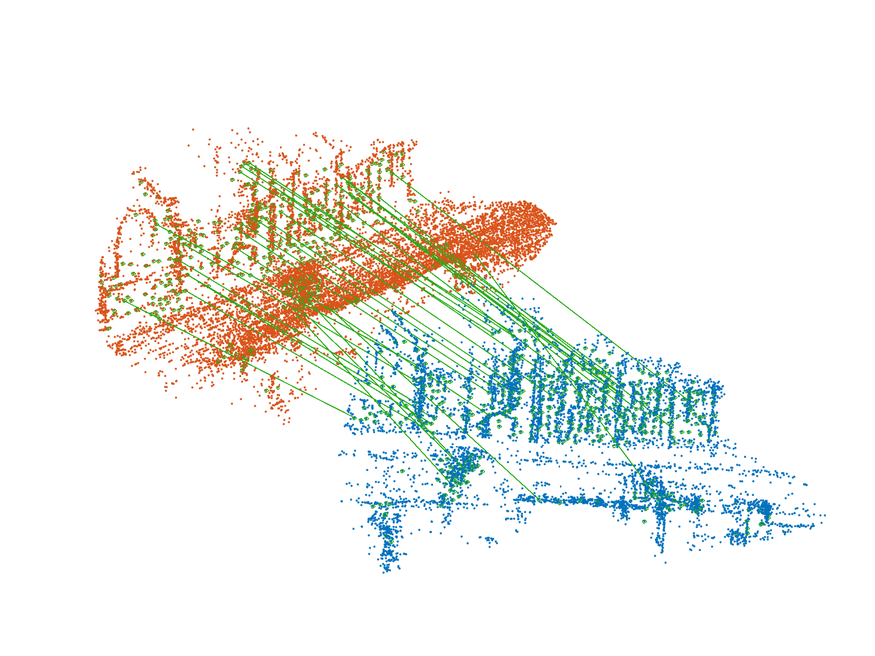}
	\includegraphics[width=0.25\textwidth, 
	height=2.2cm]{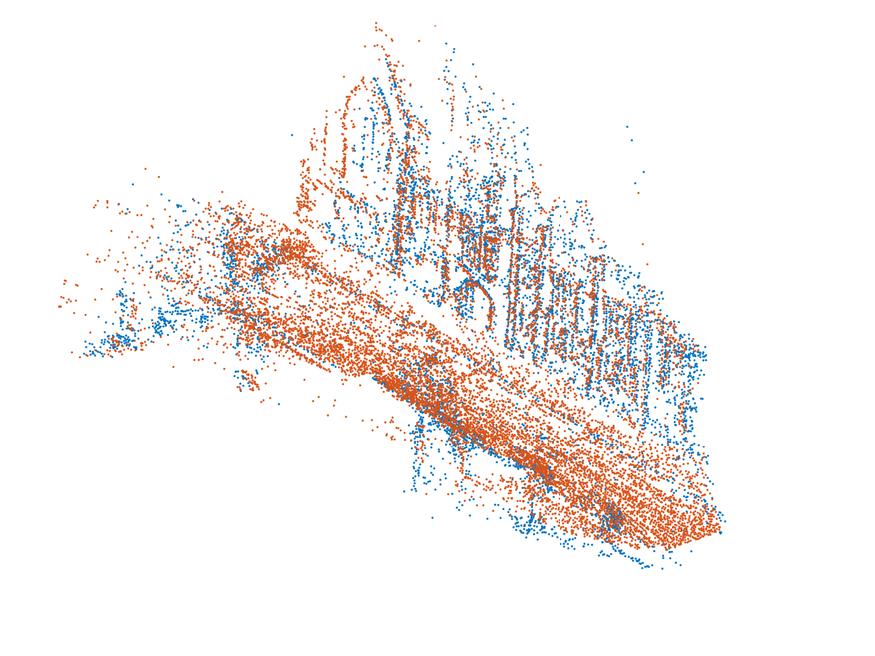}
	
	\includegraphics[width=0.22\textwidth, 
	height=2.0cm]{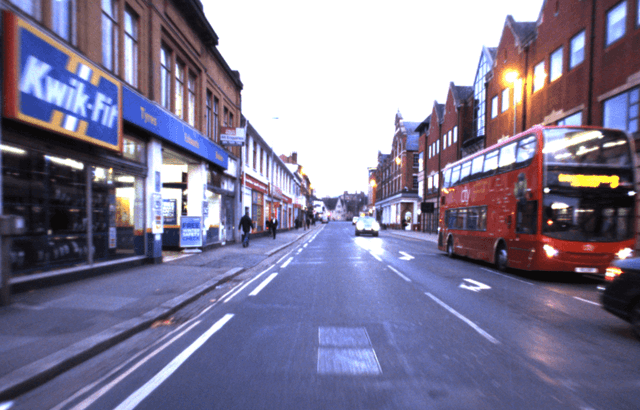}
	\includegraphics[width=0.22\textwidth, 
	height=2.0cm]{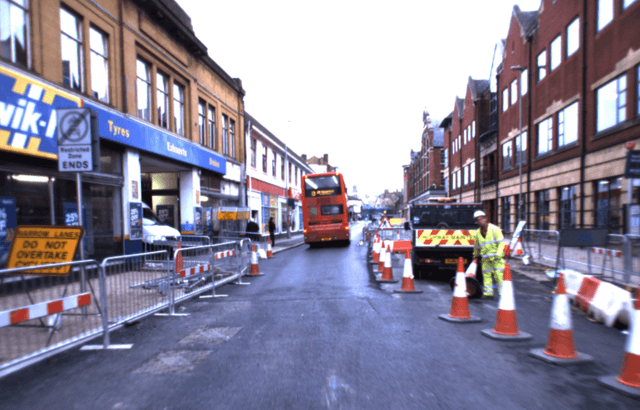}
	\includegraphics[width=0.25\textwidth, 
	height=2.2cm]{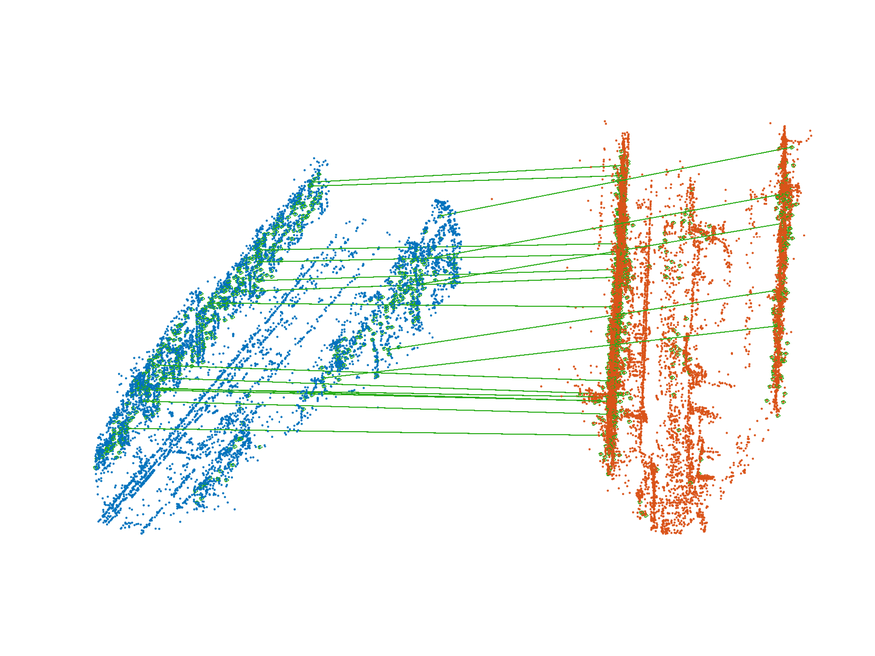}
	\includegraphics[width=0.25\textwidth, 
	height=2.2cm]{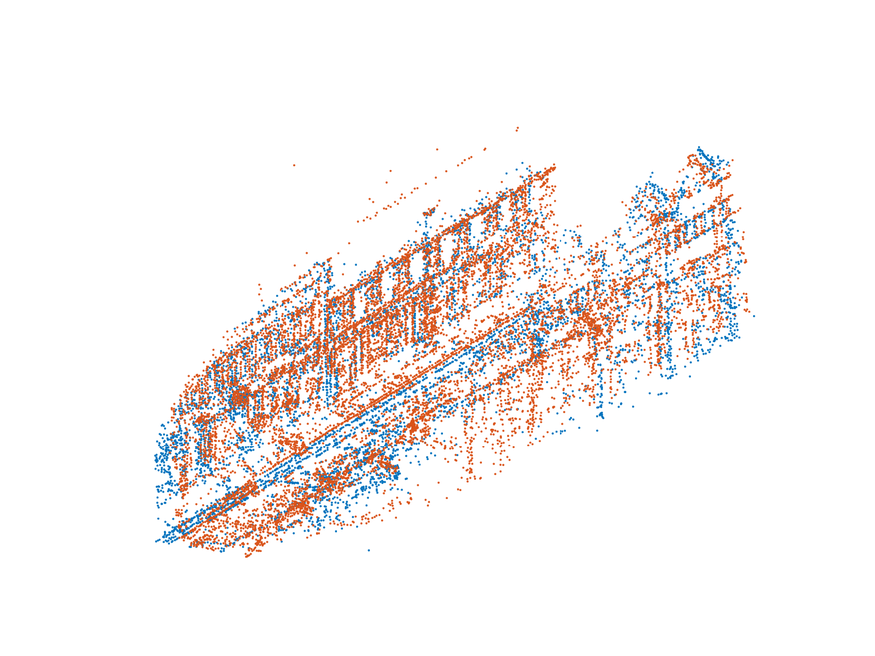}
	
	\includegraphics[width=0.22\textwidth, 
	height=2.0cm]{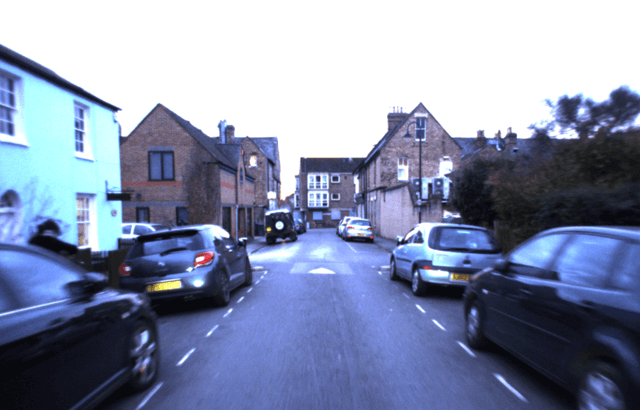}
	\includegraphics[width=0.22\textwidth, 
	height=2.0cm]{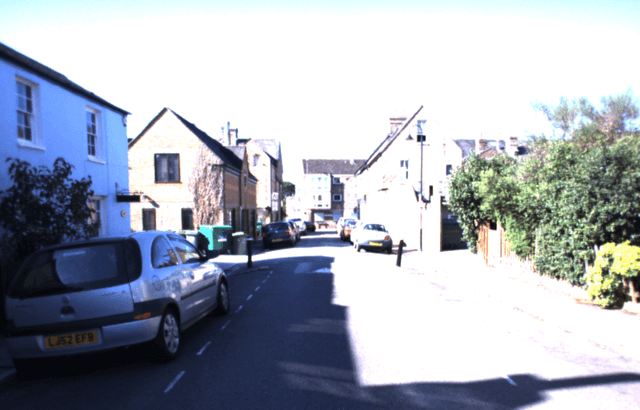}
	\includegraphics[width=0.25\textwidth, 
	height=2.2cm]{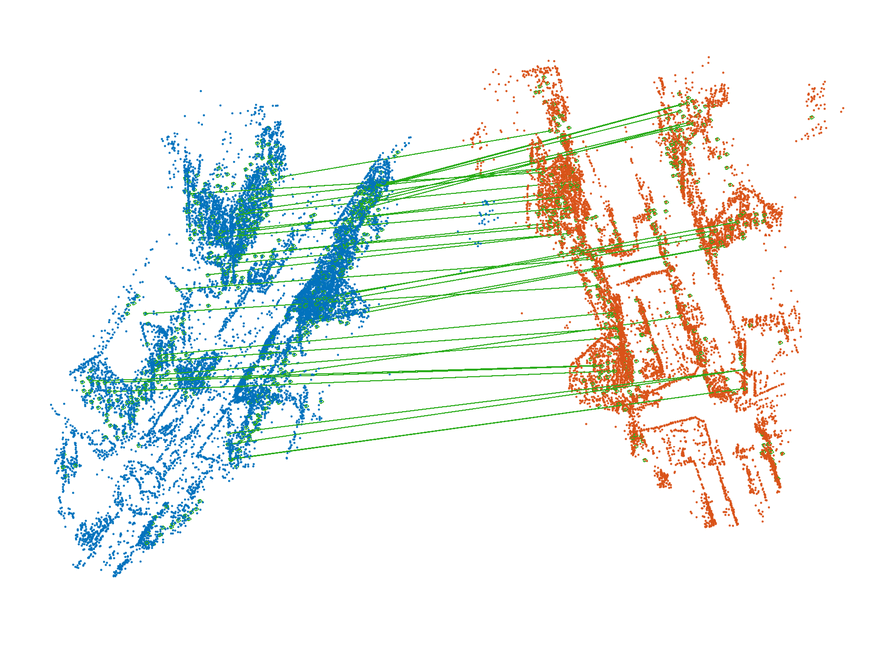}
	\includegraphics[width=0.25\textwidth, 
	height=2.2cm]{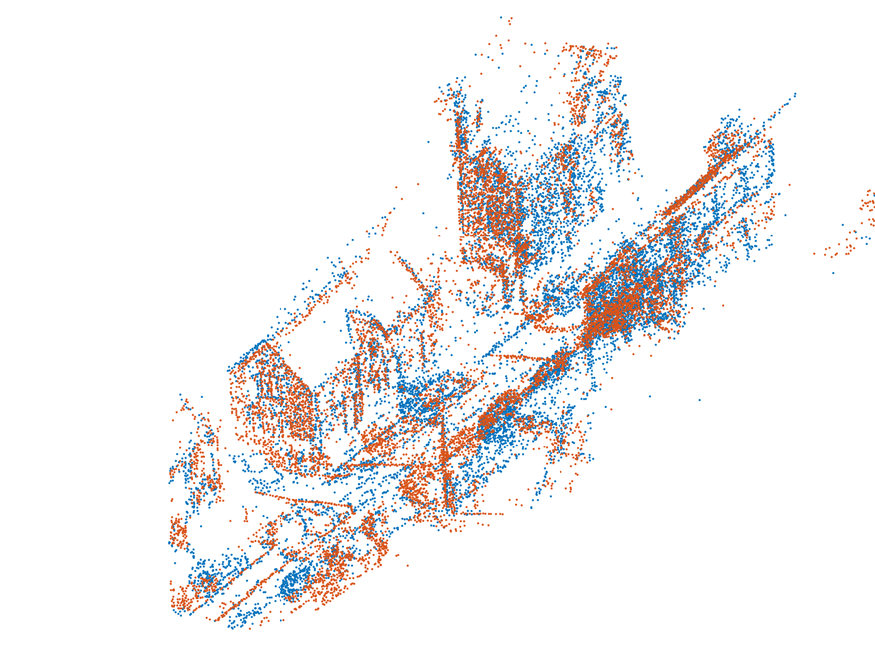}
	
	\includegraphics[width=0.22\textwidth, 
	height=2.0cm]{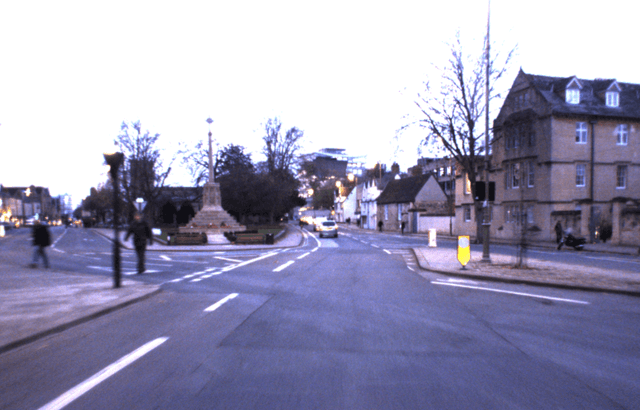}
	\includegraphics[width=0.22\textwidth, 
	height=2.0cm]{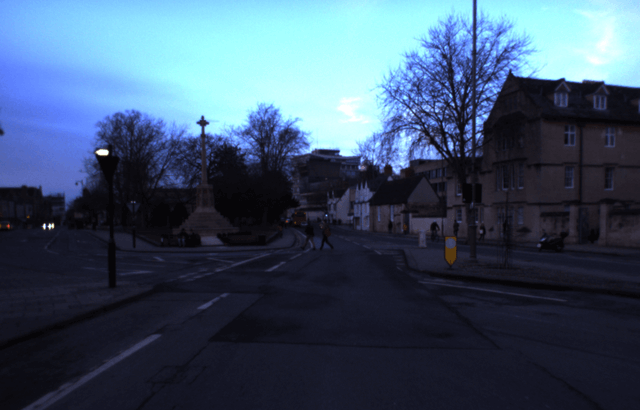}
	\includegraphics[width=0.25\textwidth, 
	height=2.2cm]{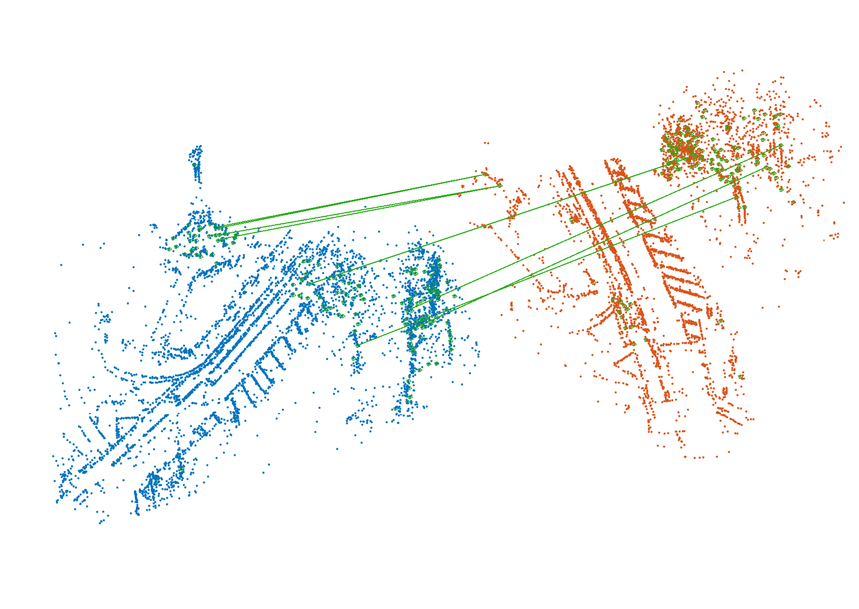}
	\includegraphics[width=0.25\textwidth, 
	height=2.2cm]{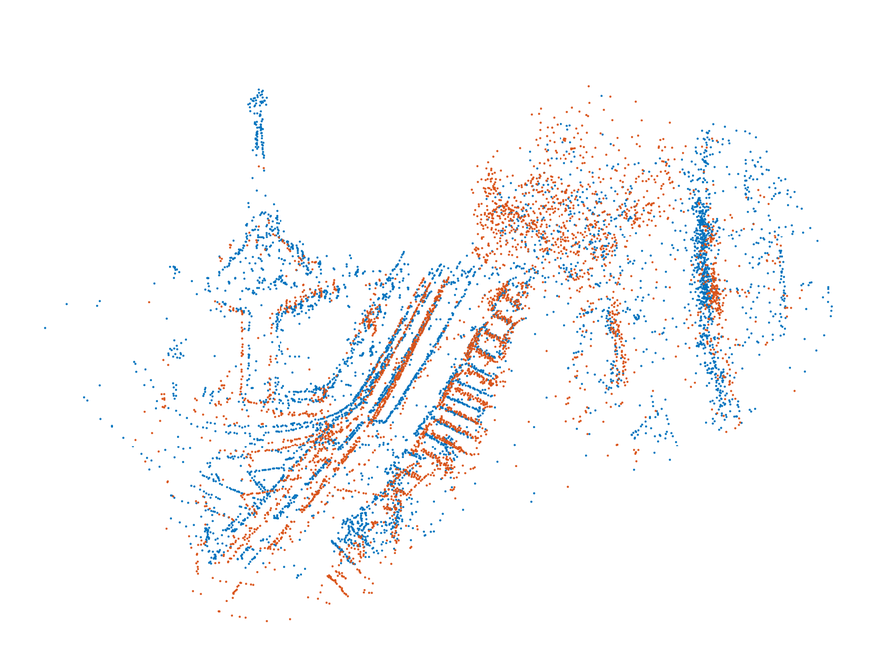}
	
	\includegraphics[width=0.22\textwidth, 
	height=2.0cm]{figures/dso_res/13_1.png}
	\includegraphics[width=0.22\textwidth, 
	height=2.0cm]{figures/dso_res/13_2.png}
	\includegraphics[width=0.25\textwidth, 
	height=2.2cm]{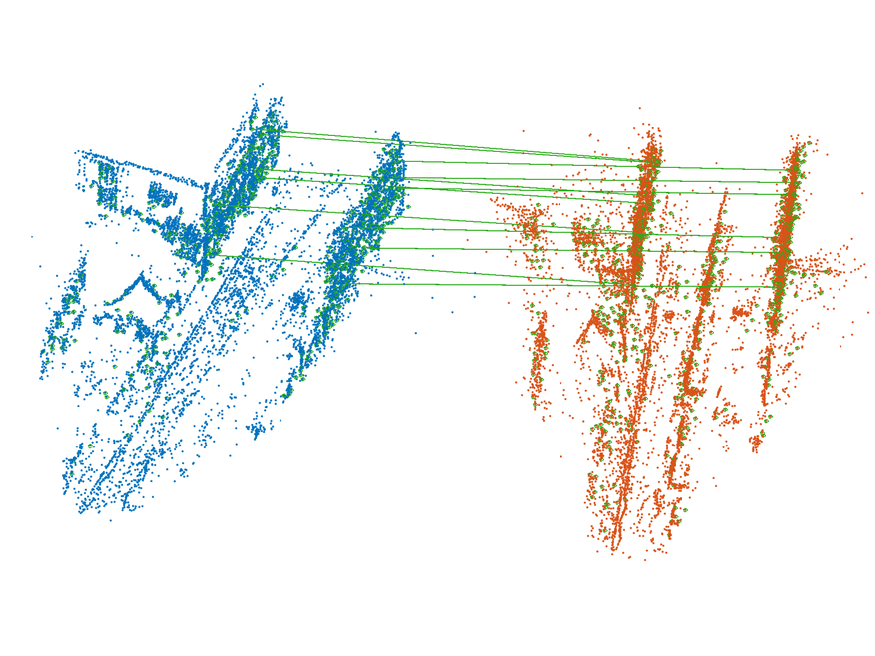}
	\includegraphics[width=0.25\textwidth, 
	height=2.2cm]{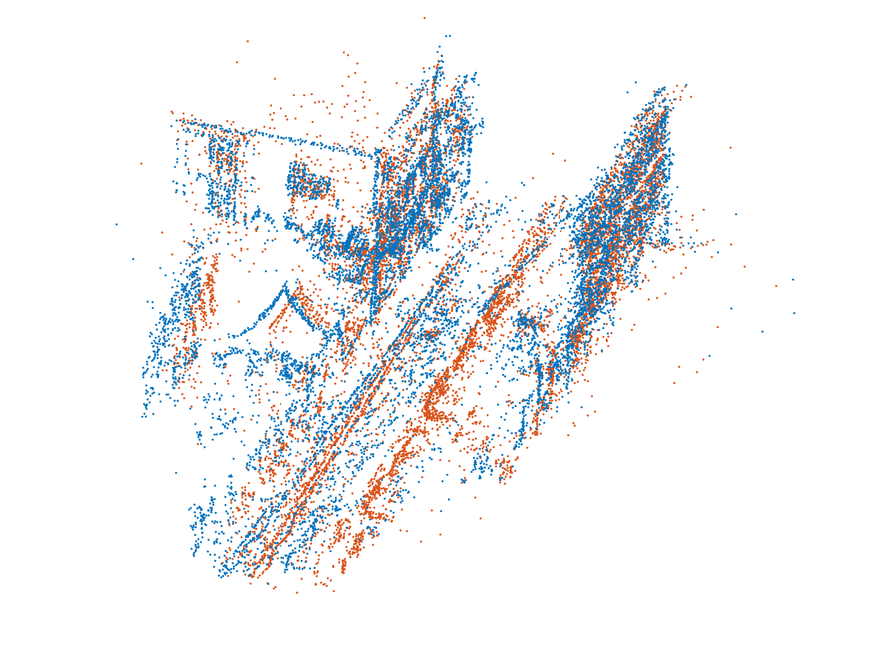}

	\caption{More results of registration of StereoDSO-Oxford. The first two columns 
	display frames from the reference and the query sequences. The last two columns show the 
	matched features (after RANSAC) and the point clouds after alignment. Note that under different 
	lighting and weather conditions, the point clouds generated by Stereo DSO have very different 
	spatial densities and distributions.}
	\label{fig:dso_vismore}
\end{figure*}

Fig.~\ref{fig:dso_vismore} shows more registration results of StereoDSO-Oxford,  which covers 
different conditions including night, dusk, direct sun, snow, rain and roadworks. Compared to LiDAR 
scanner, due to the nature of tending to reconstructed 3D points corresponding to pixels with high 
image gradients, visual SLAM methods usually generate point clouds with evidently different spatial 
densities and distributions under different lighting conditions. Such variations become even 
significant when further combined with different scene layouts. This can be clearly observed in the 
figure. Still, our keypoint detector and local descriptors trained on LiDAR points are able to 
achieve fairly good matchings without fine-tuning.

%%%%%%%%%%%%%%%%%%%%%%%%%%%%%%%%%%%%%%%%%%%%%%%%%%%%%%%%%%%%%%%%%%%%%%%%%%%%%%%%%%%%%%%%%%%%%%%%%%%%
\section{Additional Results on Global Descriptor}
%%%%%%%%%%%%%%%%%%%%%%%%%%%%%%%%%%%%%%%%%%%%%%%%%%%%%%%%%%%%%%%%%%%%%%%%%%%%%%%%%%%%%%%%%%%%%%%%%%%%

\subsection{Comparison of Different Global Aggregators}

\begin{figure}[t]
	\centering
	{\includegraphics[width=0.5\textwidth]{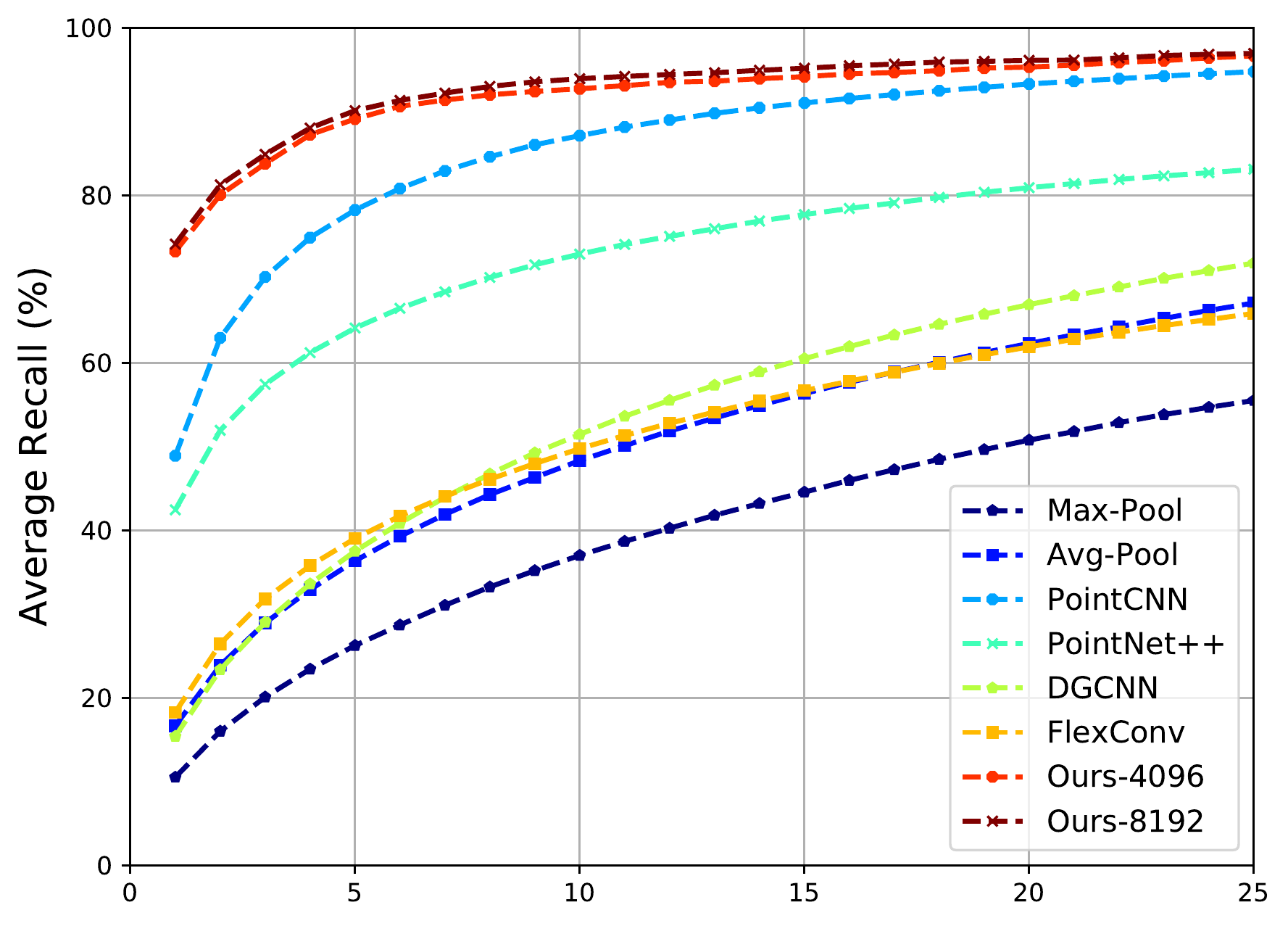}}
	\caption{Average recall of the top 25 retrievals of different approaches for assembling global 
	descriptors on the Oxford RobotCar dataset.}
	\label{fig:recall_all}
\end{figure}

As mentioned in the main paper, there exist many ways to aggregate the extracted local descriptors 
into a global one. Apart from the attention based NetVLAD model, which is adopted as our global 
assembler, we tested four other models, i.e. PointNet++\cite{pointnetplus}, Dynamic Graph CNN 
(DGCNN)~\cite{wang2019dynamic}, PointCNN~\cite{pointcnn}, and FlexConv~\cite{flex}. Max- and 
average-pooling (Max-Pool, Avg-Pool) are tested additionally as two baselines. We first explain the 
details on each model and discuss the results presented in Fig.~\ref{fig:recall_all} afterwards.

\medskip
\noindent
%-----------------------------------------------------------------------------------------------
\textbf{PoinNet++}: To construct a PoinNet++ style global assembler, we use a combination of  
Multi-Scale Grouping (MSG) layer and a Set Abstraction(SA) layer proposed in~\cite{pointnetplus}. 
The  architecture is as follows: SA(1024, [4.0, 8.0], [16, 32], [[128, 128, 256], [128, 128, 
256]]) \(\rightarrow\) SA([256, 512, 1024]) \(\rightarrow\)  FC(512, 0.5) \(\rightarrow\) FC(256, 
0.5), where SA(\(K,r,[l_1, \cdots, l_d]\)) is a SA level with \(K\) local regions of ball radius 
\(r\) and use a PointNet of \(d\) fully connected layers with width \(l_i (i = 1, \cdots, d)\). 
Note that all the models discussed in this section include a final fully connected layer to 
transform the global descriptor to a fixed dimension of 256.

\medskip
\noindent
%-----------------------------------------------------------------------------------------------
\textbf{DGCNN}: We use two consecutive EdgeConv blocks (MLP(128, 128) and MLP(256, 256)) to 
process the local descriptors, where MLP(\(a_1,..., a_n\)) is a multi-layer perceptron with the 
number of layer neurons defined as \((a_1, a_2, ..., a_n)\). The output of these two layers are 
concated which is followed by another MLP(1024) layer to generate a tensor of shape \(N \times 
1024\). Then, a global max pooling is used to get the global descriptor. 

\medskip
\noindent
%-----------------------------------------------------------------------------------------------
\textbf{PointCNN}: Inspired by PointCNN, we also test the idea of incorporating 
\goodchi-convolution 
to extract a global descriptor. A \goodchi-conv(N, C, K, D) layer takes N points as input and  
outputs a tensor of shape \(N \times C\). It is defined in a local region which constrcuted by 
sampling K input points from \(K \times D\) neighboring points, where \(D\) is the dilation rate. 
In our settings, we add four \goodchi-conv layers to transform the local features as follows: 
\goodchi-conv(2048, 256, 16, 2) \(\rightarrow\) \goodchi-conv(768, 512, 16, 2)  \(\rightarrow\) 
\goodchi-conv(384, 512,16,2 ) \(\rightarrow\) \goodchi-conv(128, 1024, 16,2)). 

\medskip
\noindent
%-----------------------------------------------------------------------------------------------
\textbf{FlexConv}: We add four additional FlexConv layers as follows, FlexConv(256, 8, 512)  
\(\rightarrow\) FlexConv(256, 8, 128) \(\rightarrow\) FlexConv(512, 8, 32) \(\rightarrow\) 
FlexConv(1024, 8, 32) , where FlexConv(\(D, k, N\)) denotes a Flex-Convolution operation on \(N\) 
input points with the neighborhood size as \(k\) and \(D\) is the dimension of the output. Then, we 
apply a Flex-Convolution at the center point \((0,0,0)\) with all the remaining points as neighbors 
to generate a global descriptor. 

\medskip
\noindent
%-----------------------------------------------------------------------------------------------
\textbf{Pooling}: The Max-Pool and Avg-Pool take the extracted local feature map as input and 
perform 
a simple max/avg pooling to produce a single global embedding. 

\medskip
\noindent
Fig.\ref{fig:recall_all} shows the average recall curves within the top 25 retrievals on the Oxford 
RobotCar dataset. We draw the overall conclusion from these results that the attention based 
NetVLAD global aggregation is so far the best option among the multiple tested models. This 
validates our choice on the global aggregator in the main paper. In addition, we can observe that 
among all the baseline methods, PointCNN far outperforms others. One possible explanation is that 
PointCNN estimates a transformation on the input points which sufficiently exploits the 3D position 
information compared to other methods. This can also account for the second highest recall obtained 
by PointNet++ which takes advantage of the 3D position implicitly by using an MSG layer. FlexConv, 
DGCNN and Avg-Pool produce similar results while Max-Pool gives the worst performance, most likely 
because a simple max pooling causes a large amount of loss of information.

\subsection{Quanlitative Results on Point Cloud Retrieval} 
\begin{figure*}
	\centering
	\includegraphics[width=0.98\textwidth]{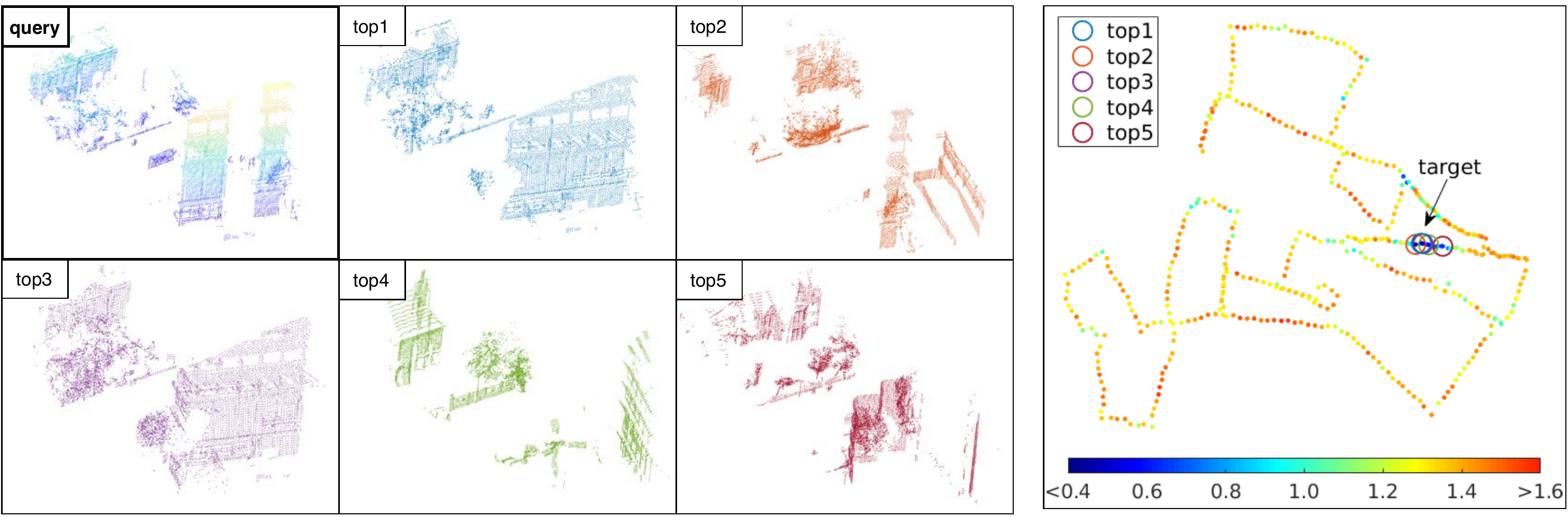}
	\includegraphics[width=0.98\textwidth]{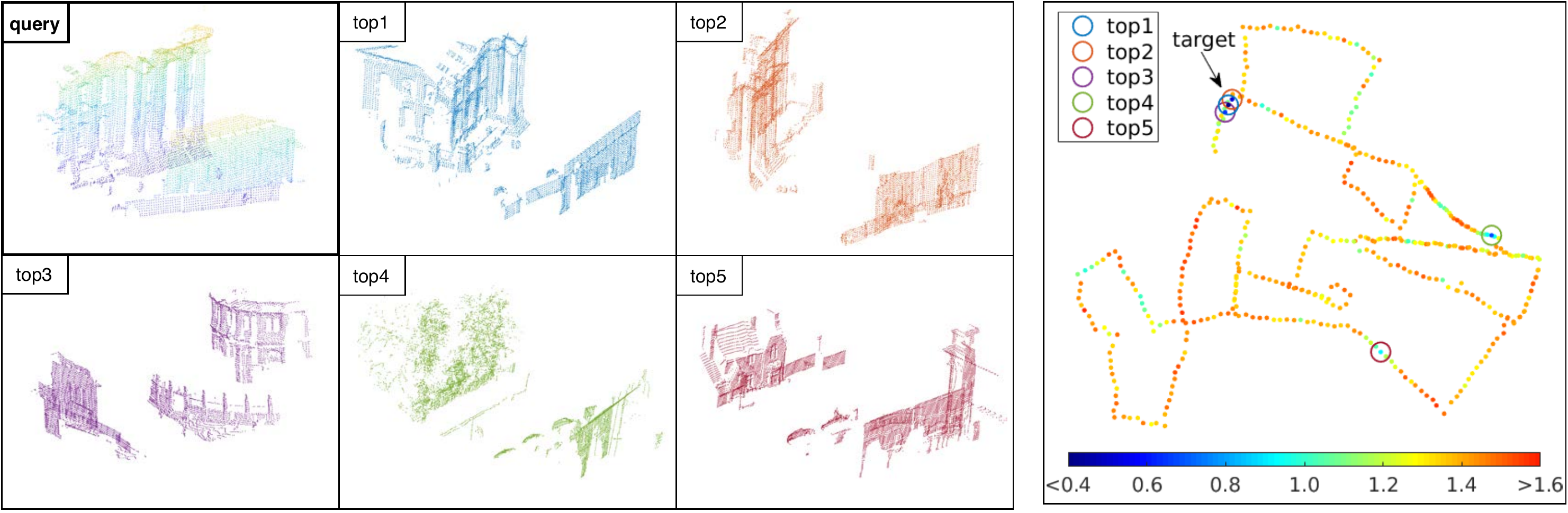}
	\includegraphics[width=0.98\textwidth]{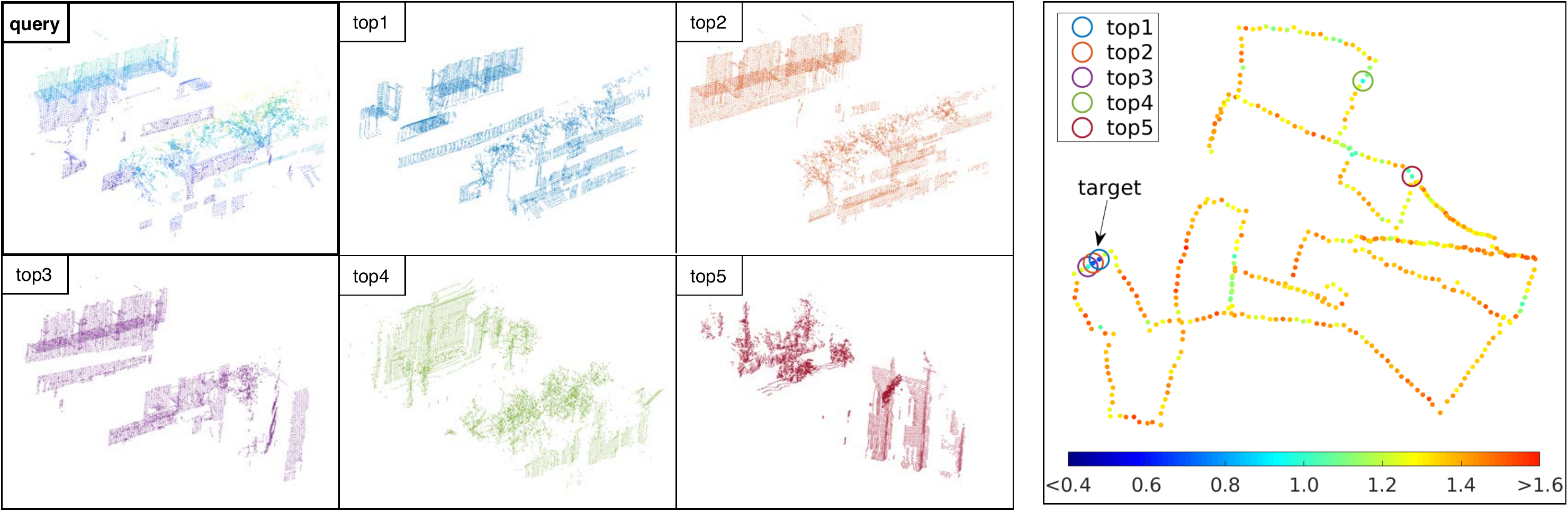}
	\caption{Visualizations of example retrieval results of our network on the Oxford RobotCar 
	dataset. For each retrieval, we display the query point cloud and the top5 matches returned 
	by our method. We also indicate the 3D locations of these point clouds in the associated 
	reference map. Colors in each query point cloud indicate different heights above ground while 
	colors of the retrieved point clouds correspond to the markers drawn in the reference maps. The 
	L2-distances between the global descriptor of the query point cloud and those of all the 
	submaps are color-coded. }\label{fig:globalvisuals}
\end{figure*}

In addition to the quantitative results of our global descriptors (Ours-8192) on point cloud 
retrieval reported in Sec. 4.3 of the main paper, we provide some qualitative examples in 
Fig.~\ref{fig:globalvisuals}. To this end, we first randomly select a full traversal (Sequence 
2014-12-10-18-10-50) as the reference map. Then three query point clouds from three other randomly 
selected traversals in the remaining 32 sequences are chosen, each representing one sample submap 
from individual testing areas as shown in the right sub-figure of Fig.~\ref{fig:split}. For each 
example, we display the query point cloud and the top5 retrieved results. In addition, the location 
of each point cloud is plotted in the reference map on the right. We use different colors to 
indicate the L2-distances between the global descriptors of the query point clouds and those of the 
submaps contained in the reference map. For each query, the best match (represented by the blue 
square) correctly overlaps with the target location (indicated by the red circle), which 
demonstrates that our global descriptors have good robustness to rotation and occlusion. It is also 
worth noting that most of the un-retrieved submaps in the reference map have relatively high 
distances in the descriptor field, demonstrating the good discriminativeness of our global 
descriptor.

%%%%%%%%%%%%%%%%%%%%%%%%%%%%%%%%%%%%%%%%%%%%%%%%%%%%%%%%%%%%%%%%%%%%%%%%%%%%%%%%%%%%%%%%%%%%%%%%%%%%
\section{Technical Details}
%%%%%%%%%%%%%%%%%%%%%%%%%%%%%%%%%%%%%%%%%%%%%%%%%%%%%%%%%%%%%%%%%%%%%%%%%%%%%%%%%%%%%%%%%%%%%%%%%%%%
\subsection{Additional Network Details}
The architectures of our local feature encoder and global descriptor assembler are illustrated 
in the main paper. Here we provide the structural details of the other two sub-networks, i.e., 
the keypoint detector and the attention map predictor.

\begin{figure}[htbp]
	\centering
	\includegraphics[width=0.7\textwidth]{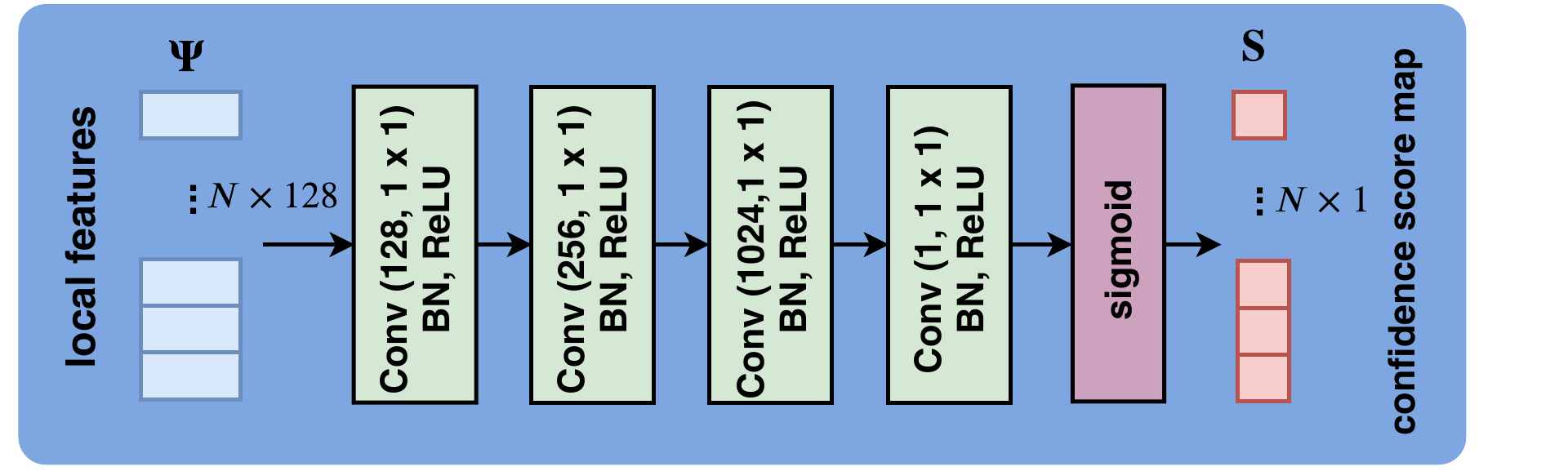}
	\caption{ The architecture of the local feature detector in 
		Fig. 2 of the main paper.}
	\label{fig:detector}
\end{figure}

\medskip
\noindent
%-----------------------------------------------------------------------------------------------
{\textbf 3D Keypoint detector.}
Fig.~\ref{fig:detector} depicts the architecture of our 3D keypoint detector. As explained in 
the main paper, it consists of 4 \(1 \times 1\) convolution layers and a sigmoid activation 
function at the end. All the convolution layers are followed by ReLU non-linear activation and 
BN (BatchNorm). The \(1 \times 1\) convolutions essentially transform each point independently 
and are invariant to point ordering. The sigmoid activation function is used to restrict the 
individual output value to \([0,1]\), which represents the matching reliability of each local 
descriptor.

\medskip
\noindent
%-----------------------------------------------------------------------------------------------
{\textbf Attention map predictor.} We use a sub-network which has similar structure as the detector 
to predict attentive weight for each local descriptor. The main component is composed of 3 \(1 
\times 1\) convolution layers with ReLU activation and BN. A softmax operation is applied to 
produce 
the final attention distribution that sums up to 1. Recall that in our case, the inputs to the 
attention predictor are the learned local descriptors, which have already encoded multi-level 
contextual information. Thus our attention predictor can benefit from this and have a relatively 
simple yet effective model structure.

\begin{figure}[htbp]
	\centering
	\includegraphics[width=0.7\textwidth]{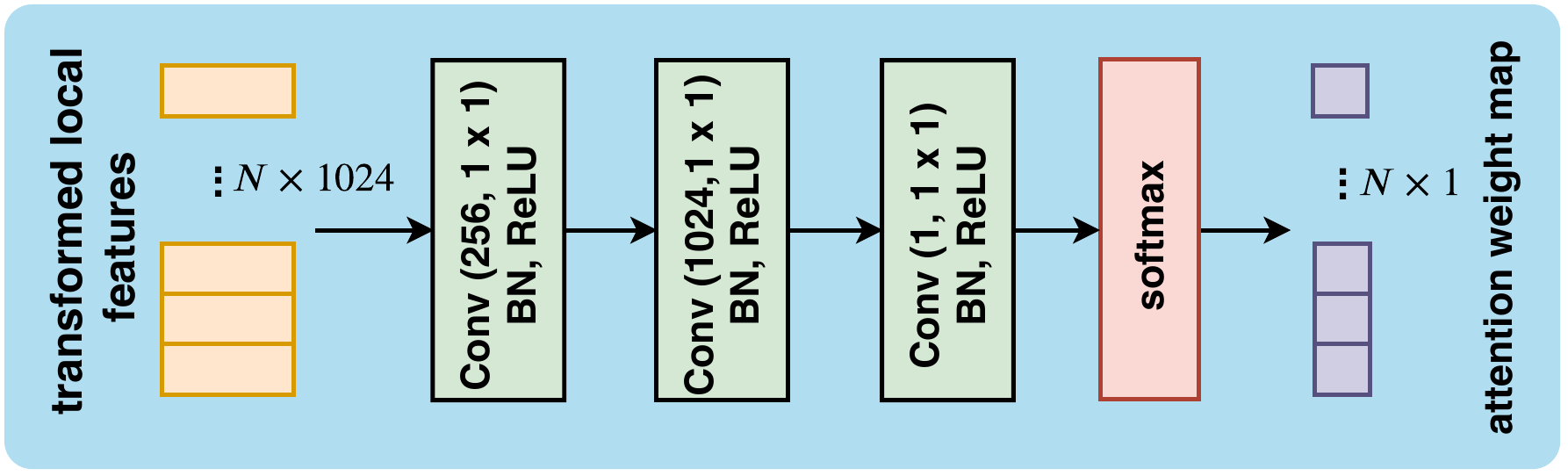}
	\caption{ The architecture of the attention predictor in Fig. 3 of the main paper.}
	\label{figattentionpredictor}
\end{figure}

\subsection{Training Data Preparation}\label{subsec:dataprep}
\begin{figure}[htbp]
	\centering
	{\includegraphics[width=0.7\textwidth]{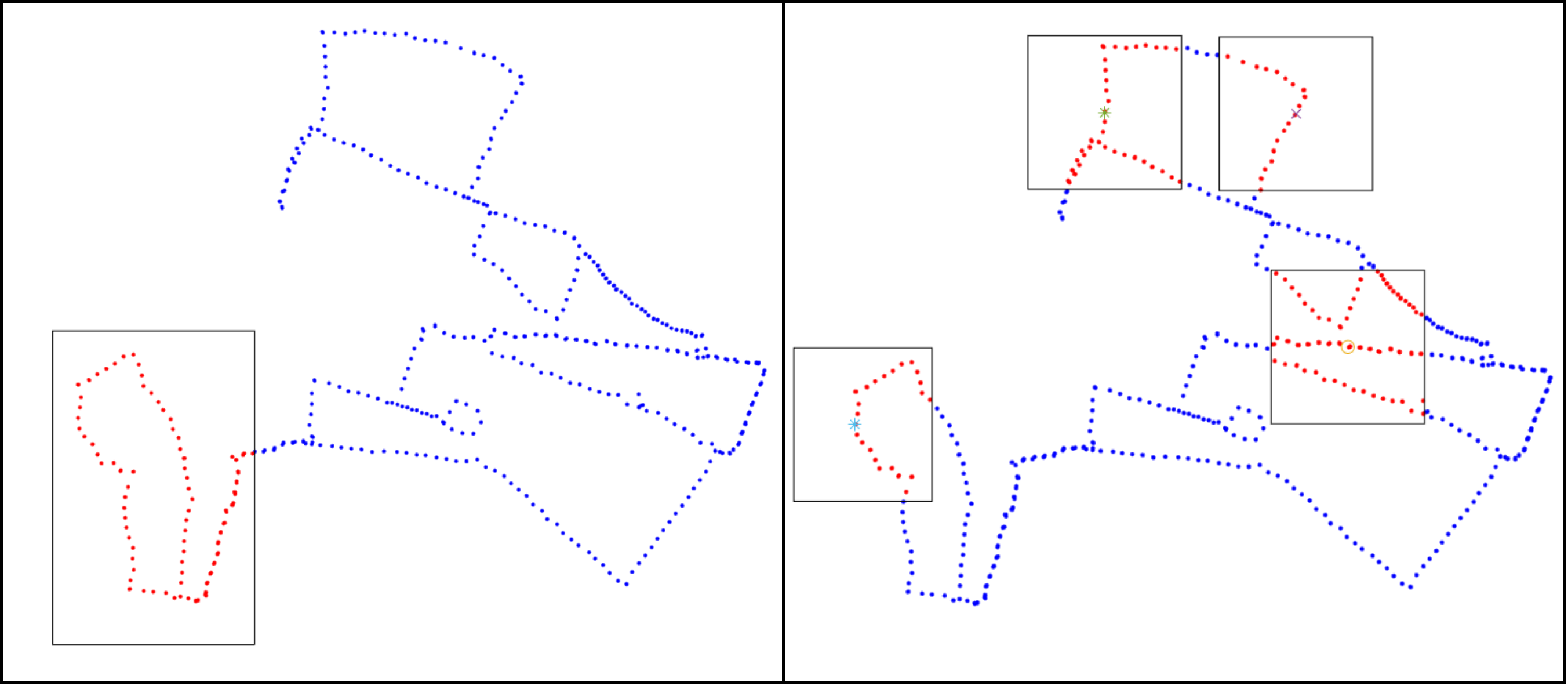}}
	\caption{Different data splitting strategies for learning local descriptors and keypoint 
	detector (left) and global descriptors (right). Blue points represent submaps in the training 
	set and red points represent those in the testing set.}
	\label{fig:split}
\end{figure}

\medskip
\noindent
%-----------------------------------------------------------------------------------------------
{\textbf Data splitting.}
We use the LiDAR points from the Oxford RobotCar dataset~\cite{robotcar} to learn both local 
and global descriptors. For fair comparisons with other methods, we split the dataset 
differently for these two tasks when constructing the training and testing data. More 
specifically, for the keypoint detection and local descriptor extraction, we follow the 
practice of 3DFeatNet~\cite{3dfeat} and make use of 40 full traversals with each traversal 
split into two disjoint sets for training and testing. The training set is constructed by the 
point clouds accumulated from the training region of the first 35 sequences. The point clouds 
collected from the testing region of the remaining 5 sequences are used to generate the testing 
set. For global descriptor learning, we adopt the split used in 
PointNetVLAD~\cite{pointnetvlad}, where a set of 44 traversals are used, both full and partial. 
Each run is then geographically split into 70\% and 30\% for training and testing, 
respectively. A comparison of the above two data splittings is illustrated in 
Fig.~\ref{fig:split}.

\medskip
\noindent
%-----------------------------------------------------------------------------------------------
{\textbf Data preprocessing.}
For each traversal, we create a 3D point cloud submap with a 20m radius for every 10m interval 
whenever good GPS/INS poses are available. According to the widely adopted convention, the 
ground planes are removed in all submaps since they are repetitive structures and 
non-informative. The resulting submaps are then downsampled using a VoxelGrid filter with a 
grid size of 0.2m. For each point cloud, 8192 points are randomly chosen on the fly during 
training. Adopting the data splitting strategies as explained above, we obtain 20,731 point 
clouds for learning local descriptor and detector and 21,026 submaps for training global 
descriptor. As mentioned in the main paper, due to the lack of accurate point-to-point 
correspondences between different sequences,  we generate training samples for local descriptor 
and detector learning with synthetic transformations (arbitrary rotations around the upright 
axis and Gaussian noise with \(\sigma_{noise} = 0.02m\)). The correspondence matrix \(\gt \in 
\realset^{N\times N}\) used in Eq. (2) and Eq. (3) in the main paper can be 
computed 
as : 
\begin{equation} \label{eq:match} 
\gt(i,j) = \mathbf{1}(\|\vp_{i}  - \mathbb{T}\vp_{j}' \|_2 < \tau), 
\end{equation} 
where \(\mathbf{1}\) is the indicator function, \(\vp_{i}\) and \(\vp_{j}'\) are the 3D centers 
of the \(i\)th and \(j\)th local patches from the anchor point cloud \(\mat{P}\) and the 
positive point cloud \(\mat{P}'\) , respectively. \(\mathbb{T}\) is the applied synthetic 
transformation. \(\tau\) is set to 0.5 in our experiments. Different from learning the local 
descriptors, the training of the global descriptor assembler can make use of the GPS/INS 
readings provided by the dataset as the supervision information. To this end, each downsampled 
submap is tagged with a UTM coordinate at its respective centroid. Similar point clouds are 
defined to be at most 10m apart and those dissimilar to be at least 50m apart.

\subsection{Two-Phase Training}
%-----------------------------------------------------------------------------------------------
In practice, we train our model in two phases to improve stability. We first focus on the local 
descriptor matching task. The input is a pair of point clouds and the output is two sets of 
matching descriptors associated with their matching confidence. The loss function used is 
formulated as: 
\begin{equation} \label{eq:loss}
L = L_{desc} + \lambda L_{det}.
\end{equation}

We use a batch size of 6 pairs of point clouds and for 
each pair of training samples, we randomly choose 512 points from the anchor point cloud as 
centers of the local patches (the points are chosen correspondingly in the transformed point 
clouds). This results in \(6 \times 512^2\) combinations for the network per batch. We use the 
Adam optimizer with a learning rate of \(10^{-4}, \beta_1 = 0.9, \beta_2 = 0.999\). The 
learning rate is successively halved every 5 epochs. In the second phase, we freeze the local 
feature encoder so that the networks are enforced to train later feature extracting layers 
which aims to model higher-level semantic information, as well as the attention map and the 
NetVLAD layer which is for aggregating local features into discriminative global features. We 
use the same lazy quadruplet loss as used in PointNetVlad~\cite{pointnetvlad} and 
PCAN~\cite{pcan} which is formulated as:

\begin{equation}
\begin{split}
\textit{L}_{lazyQuad}(\tup) &= max([\alpha + \delta_{pos} - \delta_{neg}]_{+}) \\
&+ max([\beta + \delta_{pos} - \delta_{neg*}]_{+}),
\end{split}
\end{equation}
where \( \tup\) refer to a tuple of point cloud samples in one training iteration, which 
includes an anchor point cloud \(\ancpc\), a set of positive and a negative point clouds to the 
anchor \(\{\pospc\}, \{\negpc\}\) as well as a random sample \( \mat{P}_{neg*}\) that is 
dissimilar to all the former samples. We use \(\delta_{pos/neg}\) to denote the L2-distance 
between the global descriptor vectors of \(\mat{P}_{anc}\) and \(\mat{P}_{pos/neg}\), while 
\(\delta_{neg*}\) measures the distance between \(\mat{P}_{neg*}\) and \(\mat{P}_{neg}\). The 
max operator of the first term selects the best positive in \(\{\pospc\},\) and the hardest 
negative in \(\{\negpc\}\). Similarly, the hardest negative in \(\{\negpc\}\) that gives the 
smallest \(\delta_{neg*}\) value is selected by the second loss term. \(\alpha\) and \(\beta\) 
are two different constant parameters giving the margins which are set as 0.5 and 0.2 
respectively in our experiments. In this stage, we train our model with a single batch of data 
consisting 1 \(\ancpc\), 1 \(\pc_{neg*}\), 2 \(\pospc\) and 8 \(\negpc\). The Adam optimizer is 
used with \(\beta_1 = 0.9, \beta_2 = 0.999\). The initial learning rate is set as \(5\times 
10^{-4}\) and exponentially decayed after every 10 epochs until \(10^{-5}\).The dimension of 
the output global descriptor and the number of clusters in the NetVLAD layer are set to be same 
as in ~\cite{pointnetvlad,pcan}, i.e., 256 and 64, respectively.

\subsection{Testing Set Generation by Stereo DSO }

In Sec. 4.4 of the main paper, we use the point clouds generated by Stereo DSO~\cite{stereodso} 
to evaluate the generalization capability of our method. Here we provide more details on the 
testing set used in our experiments. The Bumblebee XB3 images under the wide-baseline 
configuration from the Oxford RobotCar dataset are taken as input to Stereo DSO, which 
estimates the camera poses and pixel depths by minimizing a photometric error obtained by 
direct image alignment. To extract a local point cloud, Stereo DSO is first run on a 
sub-sequence of images and all the valid 3D points associated with each keyframe are collected. 
These points are then projected into one common coordinate system utilizing the estimated 
relative camera poses.

To cover a wide range of day-time and weather conditions, eight sequences are chosen as listed 
in Tab.~\ref{tab:sequenceinfo}. One sequence in the overcast condition is set as the reference 
traversal, where the Stereo DSO is run on the full sequence. For each of the other seven 
sequences, 50 different images are uniformly selected as the starting frames and the 
corresponding local point clouds are extracted using the method described above. After 
filtering out invalid ones due to failures of Stereo DSO under some challenging situations, we 
obtain 332 point clouds as shown in Tab.~\ref{tab:sequenceinfo}. To generate the ground truth 
transformations, the corresponding point clouds accumulated from LiDAR scans are also collected 
to calculate the accurate relative poses by performing ICP between overlapped point clouds. 
This gives us 318 pairwise poses for testing. We crop each extracted point cloud with a fixed 
radius of 30m around its centroid followed by a downsampling step with a VoxelGrid filter with 
the fixed grid size of 0.2m. Lastly, we randomly rotate each point cloud around the vertical 
axis to evaluate the rotational invariance. We thank the authors of Stereo DSO again for their 
great help on generating the point clouds for us. These point clouds as well as the related 
data splits explained in Sec.~\ref{subsec:dataprep} will be released to the community upon the 
publication of this paper.

\begin{table}[htbp]
	\footnotesize
	\centering
	
	\begin{tabular}{lcc}
		\toprule        
		Sequence & Conditions  & Valid point clouds   \\
		\midrule
		2014-12-02-15-30-08 (ref) & overcast & 107 \\
		\midrule
		2014-11-14-16-34-33 & night & 8 \\
		2014-12-16-18-44-24 & night & 17 \\
		2014-11-25-09-18-32 & rain & 22 \\
		2015-02-03-08-45-10 & snow & 39 \\
		2015-03-24-13-47-33 & sunny & 43 \\
		2015-02-20-16-34-06 & dusk, roadworks & 47 \\
		2015-11-10-14-15-57 & overcast, roadworks & 49 \\
		\bottomrule
	\end{tabular}
	\caption{Number of point clouds generated by Stereo DSO for point cloud registration 
	evaluation. 
	}
	\label{tab:sequenceinfo}
\end{table}

\subsection{Weak Supervision Used in the Ablation Study}
In the ablation study, we compare our local descriptors with those trained using a weak 
supervised manner proposed by 3DFeatNet~\cite{3dfeat}. The weak supervision information is 
provided at the submap-level which is based on the ground truth coarse registrations of the 
point clouds in the training set. More specifically, . the input to the network in each 
iteration is an anchor point cloud \(\ancpc\) with a positive  point cloud \(\pospc\) and a 
negative point cloud \(\negpc\) to the anchor. The model is trained using the triplet loss 
computed by :

\begin{equation}
\textit{L} = \sum_{n = 1}^{N}
\left[ \min_{\vec{x}_i \in \goodchi_{pos}}
\|\vec{x}_n - \vec{x}_i\|_2
-  \min_{\vec{x}_j \in \goodchi_{neg} } \|\vec{x}_n - \vec{x}_j\|_2
+ \gamma \right]_{+},
\end{equation}

where \(N\) is the number of local clusters extracted in each point cloud. Each local descriptor 
\(\vec{x}_n\) in \(\ancpc\) can find its closest descriptor in both \(\pospc\) and \( \negpc\). The 
former is considered as the correct local match and the latter one is considered as the hardest 
negative. \(\gamma\) is the margin which is set as 0.2 in our experiments.

\clearpage
\bibliographystyle{splncs04}
\bibliography{egbib}